\documentclass{article}

\PassOptionsToPackage{numbers, compress}{natbib}

\usepackage[preprint]{vrh_preprint}

\usepackage[utf8]{inputenc} %
\usepackage[T1]{fontenc}    %
\usepackage[hidelinks]{hyperref}       %
\usepackage{url}            %
\usepackage{booktabs}       %
\usepackage{amsfonts}       %
\usepackage{amsmath}        %
\usepackage{amssymb}        %
\usepackage{nicefrac}       %
\usepackage{microtype}      %
\usepackage{xcolor}         %
\usepackage{graphicx}       %
\usepackage{enumitem}       %
\usepackage{multirow}       %
\usepackage{bbm}
\usepackage{caption}
\usepackage{tabularx}
\usepackage{pifont}
\usepackage{wrapfig}
\newcommand{\cmark}{\ding{51}} %
\newcommand{\xmark}{\ding{55}} %
\usepackage[percent]{overpic}
\usepackage{subcaption}
\usepackage{placeins}

\newcommand{\Ours}{VRH}

\title{\textit{Retrieval Heads Meet Vision}: Uncovering How VLMs Locate and Extract Visual Information}

\author{Chanho Park\textsuperscript{$1$}  $\quad$ Daehyeon Choi\textsuperscript{$1$}  $\quad$ Jihyun Lee\textsuperscript{$2$} $\quad$ Minhyuk Sung\textsuperscript{$1$} 
\\\textsuperscript{$1$}KAIST $\quad$ \textsuperscript{$2$}Independent Researcher
\\{\tt\small \{charlieppark, daehyeonchoi, mhsung\}@kaist.ac.kr}
\\{\tt\small jihyunlee1027@gmail.com}
}

\begin{document}
\maketitle

\begin{figure}[!h]
  \centering

    \includegraphics[width=\linewidth]{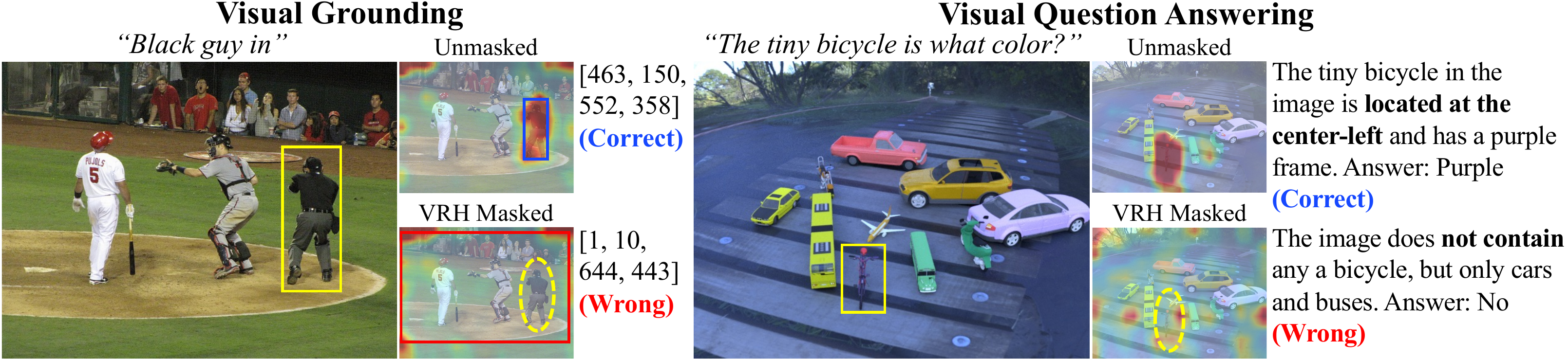}
    \caption{
    \textbf{Visual Retrieval Heads mediate visual reference resolution in VLMs.}
We identify a sparse set of attention heads, \emph{Visual Retrieval Heads} (VRHs), that are causally necessary for resolving the image region referred to by the text prompt.
On visual grounding (\textbf{left}), masking the VRHs redirects prediction-stage attention away from the ground-truth referent and causes the model to output a delocalized bounding box.
The same heads, discovered only from grounding, also transfer to visual question answering (\textbf{right}): masking them removes attention to the referred object and leads to a fluent but visually unfaithful answer.
Yellow boxes denote ground-truth referents; dashed yellow ellipses indicate regions where attention vanishes after VRH masking.
    }
  \label{fig:teaser}
\end{figure}
\vspace{0.5\baselineskip}

\begin{abstract}
Vision-language models (VLMs) can locate an image region referred to by a text prompt and route the corresponding visual evidence to the output, yet the internal mechanism behind this behavior is not understood. Inspired by retrieval heads in large language models, we ask whether VLMs contain an analogous mechanism for visual retrieval. We answer affirmatively by introducing \emph{Visual Retrieval Heads} (VRHs), a small subset of attention heads (about 1.7--2.6\%) that are causally responsible for grounding text descriptions to image regions. To find them, we recast existing head-scoring methods under a unified design space over query tokens, key aggregation, and cross-sample aggregation. We then show that scoring attention from output prediction tokens with a sum over the ground-truth referent region most reliably identifies causal heads. Across eleven VLMs and five referring-expression benchmarks, masking only the top 20 VRHs reduces grounding accuracy by up to 80 percentage points, while masking the same number of random heads has little effect. Beyond replicating the causal-sparse-universal triad established for text retrieval heads, VRHs exhibit several properties not previously reported: they \emph{generalize across visual reference tasks}, remaining causal on attribute, spatial, counting, and visual-math benchmarks despite being discovered through bounding-box prediction; they are \emph{functionally specific}, preserving output format while corrupting localization; and they are \emph{architecturally shared}, transferring causally across VLMs that share an LLM backbone but differ in vision encoder, projector, and instruction tuning.
\end{abstract}

\section{Introduction}

\vspace{-0.2\baselineskip}

How do foundation models locate a specific piece of information buried in a long input context? This ``needle-in-a-haystack'' question has recently been answered, at least partially, for large language models (LLMs): a small set of attention heads called \emph{retrieval heads}~\citep{wu2025retrieval} is largely responsible for routing relevant input tokens to the output. These heads exhibit three properties that together make them a meaningful unit of analysis. They are \emph{causal}: ablating them collapses long-context retrieval while leaving fluency intact. They are \emph{sparse}: fewer than 5\% of all heads carry the behavior. And they are \emph{universal}: they appear across model families and scales, suggesting that long-context retrieval converges onto a shared internal mechanism rather than an idiosyncratic implementation.

This observation motivates our key question: whether \emph{Vision-Language Models (VLMs)} contain an analogous mechanism for \emph{visual} retrieval---locating an image region referred to by a text prompt and routing it to the output.
Existing extensions of retrieval-head analysis to VLMs do not yet answer it. OCR Heads~\citep{baek2025ocrhead} and VERA~\citep{pei2026vera} identify heads that retrieve textual evidence from images, but only when the image is itself a rendering of text; the retrieval target is still a string of characters, not a general visual concept. Localization Heads~\citep{kang2024locheads} move beyond rendered text and aim to identify heads whose attention maps align with referred objects, but the analysis stops at attention shape: the heads are \emph{not} characterized by causal necessity.

In this work we ask whether causal, universal, and sparse retrieval heads exist in VLMs for general image retrieval, and how to find them. We answer affirmatively by introducing \emph{Visual Retrieval Heads} (VRHs), a small subset of attention heads that are causally responsible for grounding text descriptions to image regions. We show that these heads share the defining properties of their text-only counterparts.

We use referring-expression visual grounding~\citep{yu2016refcoco} as our probing task: given an image and a referring expression, the model must output a bounding box for the referent. This task makes the retrieval target spatially explicit; the ground-truth bounding box defines exactly which visual tokens the model needs to consult. This lets us define a head's contribution to retrieval by how strongly it attends from the model's prediction-stage tokens to the visual tokens inside that box.

Beyond identifying VRHs, our contributions are three-fold. First, on the \emph{methodology} side, we observe that existing head-scoring methods can all be written in a single form, with a per-head score obtained by aggregating attention weights over a set of \emph{key tokens} (the tokens to be retrieved) for each \emph{query token}, and then aggregating these per-query scores across all queries. Under this view, prior methods differ along three axes: the query token set (input prompt tokens vs.\ output prediction tokens), the key aggregation (an argmax indicator for whether the most-attended key lies in the retrieval
region vs.\ a summation of attention weights over the region), and the cross-sample aggregation (averaging scores across samples vs.\ voting). For visual grounding, we let the key tokens be the visual tokens that overlap the ground-truth bounding box, and we search this design space directly, scoring each variant by the drop in grounding accuracy when its top-ranked heads are masked. We find that the choice of key aggregation and cross-sample aggregation matters little, but the choice of query tokens matters a great deal: scoring at the \emph{output} prediction stage with a sum over the retrieval region consistently identifies the most causally important heads. We adopt it as our default scoring function.

Second, on the \emph{empirical} side, we demonstrate the causality, sparsity, and universality of the VRHs identified by our scoring function through comprehensive experiments across eleven VLMs built on eight language-model backbones~\citep{bai2025qwen25vl,bai2025qwen3vl,zhu2025internvl3,wang2025internvl35,yang2025magma,li2024llavanext,steiner2024paligemma2,wu2024deepseekvl2,chen2024internvl25,glm2025glm4v,abouelenin2025phi4rv} and five grounding benchmarks (RefCOCO/+/g~\citep{yu2016refcoco,mao2016refcocog}, RefSpatialBench~\citep{zhou2025roborefer}, Toloka~\citep{ustalov2023toloka}). Masking the top 20 VRHs---about 1.7--2.6\% of all heads---reduces grounding accuracy by up to 80 percentage points, while masking the same number of random heads has negligible effect, demonstrating their causal role.

Third, most importantly, our \emph{characterization} of VRHs reveals four properties that distinguish them from previously studied heads. (i)~VRHs \emph{generalize across visual reference tasks}: although we discover them through bounding-box prediction, they remain causal on attribute, spatial reasoning, counting, and visual math benchmarks (VAW, Spatial457, SpatialRGPT, CV-Bench, MMStar, CountBenchQA, MathVista)~\citep{pham2021vaw, wang2025spatial457, cheng2024spatialrgpt, tong2024cvbench, chen2024mmstar, paiss2023countbench, lu2023mathvista}, suggesting that the mechanism they implement is visual reference resolution in general, not object grounding in particular. (ii)~VRHs are \emph{functionally specific}: masking them preserves output-format validity but the predicted box is delocalized, the visual analogue of the ``fluent-but-not-factual'' failure mode reported for text retrieval heads. (iii)~VRHs are \emph{architecturally shared}: VRHs found in one VLM remain causally critical when masked in another VLM that shares the same LLM backbone but differs in vision encoder, projector, and instruction-tuning recipe. (iv)~VRHs are \emph{stably detectable}: results are consistent across output-token choices and scoring samples, with as few as 5 to 10 grounding examples recovering nearly the same top-ranked set as the full 200-example set.

\vspace{-0.4\baselineskip}

\section{Related Work}
\label{sec:related}

We discuss prior works on functionally specialized attention heads in large language models (LLMs) and vision-language models (VLMs).
Specifically, we focus on methods for identifying heads specialized for \textit{retrieval}-related tasks, which are most closely related to our setting.

Most existing methods in this domain identify such specialized heads by quantifying the importance of each head based on inference-time attention patterns~\cite{wu2025retrieval,zhang2025qrhead,baek2025ocrhead, pei2026vera}.
A pioneering work along this direction is Retrieval Heads~\cite{wu2025retrieval}, which detects heads specialized for text retrieval in LLMs by measuring how often the output answer token most attends to the correct input token to be retrieved, i.e., the \textit{needle} token.
QRHead~\cite{zhang2025qrhead} is a follow-up work showing that analyzing attention patterns from input prompt tokens (\textit{cf.} output tokens in Retrieval Heads) to the input retrieval source tokens is more effective for detecting specialized heads in real-world scenarios.

Beyond text-only retrieval in LLMs, OCR Heads~\cite{baek2025ocrhead} is one of the earliest works to extend the head-identification method of Retrieval Heads to vision-language models (VLMs), discovering specialized heads for optical character recognition.
Similarly, VERA~\cite{pei2026vera} identifies VLM heads specialized for long-context visual document retrieval.
However, both of these VLM-based specialized heads are limited to \textit{visual text documents} as the input retrieval source. 
The first work on specialized heads for visual retrieval from \textit{general images} is Localization Heads~\cite{kang2024locheads}, but it does not verify the causality of the detected heads, i.e., their causal effect on retrieval performance, which is a key property of retrieval specialized heads emphasized in all the aforementioned works~\cite{wu2025retrieval,zhang2025qrhead,baek2025ocrhead,pei2026vera}.

In Tab.~\ref{tab:prior_head_comparison}, we summarize the characteristics of these prior works.
To the best of our knowledge, our work is the first to identify a set of attention heads specialized for visual retrieval from \textit{general images} while exhibiting \textit{causality} in the model's visual retrieval performance. Note that we further discuss and compare the methodological differences among these prior works in Sec.~\ref{subsec:unified_form}.

\vspace{-0.5\baselineskip}

\begin{table}[!h]
\centering
\small
\caption{\textbf{Comparison of prior works on retrieval-related specialized heads.}}
\label{tab:prior_head_comparison}
\vspace{0.5\baselineskip}
\resizebox{\linewidth}{!}{\begin{tabular}{lcccccc}
\toprule
 & Retrieval Heads~\cite{wu2025retrieval} & QRHead~\cite{zhang2025qrhead} & OCR Heads~\cite{baek2025ocrhead} & VERA~\cite{pei2026vera} & Loc. Heads~\cite{kang2024locheads} & \textbf{\Ours{} (Ours)} \\
\midrule
Input Source 
& Text 
& Text 
& Visual document 
& Visual document 
& General image 
& General image \\
Causality 
& \cmark  
& \cmark  
& \cmark  
& \cmark  
& \xmark 
& \cmark  \\
\bottomrule
\end{tabular}
}
\end{table}
\vspace{-0.5\baselineskip}

\section{Toward Identifying Visual Retrieval Heads}
\label{sec:method}

We introduce our method for identifying \textit{Visual Retrieval Heads} (VRHs), a subset of attention heads within vision-language models (VLMs) specialized for retrieving visual information relevant to an input text prompt.
Our detected VRHs exhibit important desired properties of functionally specialized heads: (1) \textit{causality}: VRHs are causally responsible for visual retrieval performance, such that masking these heads significantly degrades the VLM's retrieval behavior; (2) \textit{universality}: VRHs can be found across multiple VLM families and benchmarks; and (3) \textit{sparsity}: VRHs constitute only a small subset of attention heads, revealing that visual retrieval behavior in VLMs is concentrated in a few heads.
To the best of our knowledge, ours is the first work to discover specialized heads in VLMs that are causally responsible for visual retrieval from general images.

In what follows, we first introduce our setup, where visual grounding is used as our probing task (Sec.~\ref{sec:preliminary}).
We then show that existing \textit{head-scoring methods}, i.e., methods for quantifying the importance of individual heads for a target functionality, can be expressed under a unified formulation.
This formulation allows us to systematically explore the design space of scoring methods for detecting VRHs (Sec.~\ref{subsec:unified_form}).
Based on this analysis, we propose our VRH scoring method by holistically examining different scoring method variants through causal validation, \textit{i.e.}, by measuring how much visual grounding accuracy degrades when the detected heads are masked (Sec.~\ref{subsec:causal_validation}).

\subsection{Visual Grounding as Probing Task}
\label{sec:preliminary}

We use visual grounding as a probing task for analyzing the visual retrieval behavior of VLMs and detecting VRHs.
In this task, a VLM is asked to localize the object in the input image referred to by the input text prompt.
More specifically, each sample in a visual grounding dataset is represented as $s = (I_s, q_s, B_s)$, where $I_s$ is an image, $q_s$ is a referring expression, and $B_s$ is the ground-truth bounding box of the referred object.
Given $(I_s, q_s)$, the VLM generates a text token sequence that is parsed into a predicted box $\hat{B}_s$.
A prediction is judged correct if its intersection-over-union (IoU) with the ground-truth box exceeds a threshold,
$
  \mathrm{IoU}(\hat{B}_s, B_s) \ge \tau,
  \label{eq:iou_correct}
$
where $\tau$ is typically set to $0.5$.

\subsection{Unified Formulation of Head-Scoring Methods}
\label{subsec:unified_form}

To effectively design a method to detect VRHs responsible for this visual grounding task, we first review prior works on detecting functionally specialized heads in other retrieval-related tasks in LLMs and VLMs (e.g., text retrieval~\cite{wu2025retrieval,zhang2025qrhead}, optical character recognition~\cite{baek2025ocrhead,pei2026vera}). 
The key idea in these works is to introduce a \textit{head-scoring method} to quantify the importance of each attention head for the target task and detect the specialized heads based on it. 
Here, the importance of each head is typically scored based on its attention weights computed between: (1) \textit{a set of query tokens} ($\mathcal{Q}_s$) where retrieval behavior is expected to occur, and (2) \textit{a set of key tokens} ($\mathcal{K}_s$) that correspond to the correct evidence region to be retrieved from the input. In our visual grounding setting, this evidence region is the annotated target region; therefore, we instantiate $\mathcal{K}_s$ as the set of visual tokens whose patches overlap with the annotated bounding box $B_s$.

For example, existing text retrieval head detection methods in LLMs~\citep{wu2025retrieval,zhang2025qrhead} assign high importance to heads that place strong attention from either \textit{output tokens}~\citep{wu2025retrieval} (which express the retrieval prediction) or \textit{input prompt tokens}~\citep{zhang2025qrhead} (which specify what should be retrieved) onto input tokens corresponding to the correct retrieval region.

More specifically, let $\bar{\Theta}_{l,h}$ denote the importance score of head $h$ at layer $l$, computed by aggregating the per-sample head importance scores $\Theta_{l,h}(s)$ over samples $s \in \mathcal{S}$ in the scoring dataset.
In most of the existing methods~\cite{wu2025retrieval,zhang2025qrhead,baek2025ocrhead,pei2026vera}, both the final head score $\bar{\Theta}_{l,h}$ and the per-sample head score $\Theta_{l,h}(s)$ can be expressed under the following unified formulations:
\begin{equation}
\begin{alignedat}{2}
  \bar{\Theta}_{l,h}
  &=
  \Omega_{s \in \mathcal{S}}
  \underbrace{\left[
    \Theta_{l,h}(s)
  \right]}_{\text{per-sample score}},
  \quad 
  \Theta_{l,h}(s)
  &=
  \frac{1}{|\mathcal{Q}_s|}
  \sum_{t \in \mathcal{Q}_s}
  \Phi_{j \in \mathcal{K}_s}
  \left[
    \alpha_{l,h}(t,j)
  \right],
\end{alignedat}
\label{eq:head_scoring_template}
\end{equation}

\vspace{-0.2\baselineskip}

\noindent where $\Omega_{s \in \mathcal{S}}$ denotes a cross-sample aggregation function that aggregates the per-sample scores $\Theta_{l,h}(s)$ over all samples $s \in \mathcal{S}$.
In the above equation, each per-sample score $\Theta_{l,h}(s)$ is computed by summarizing the attention weights $\alpha_{l,h}(t,j)$, which reflect how strongly head $(l,h)$ attends from a set of query tokens $t \in \mathcal{Q}_{s}$ to a set of key tokens $j \in \mathcal{K}_{s}$.
To this end, the key aggregation function $\Phi_{j \in \mathcal{K}_s}$ first aggregates the attention weights over the set of key tokens into a scalar per-query score, and these per-query scores are then averaged over all selected query tokens to finally obtain the per-sample score $\Theta_{l,h}(s)$.

Under this unified formulation, existing methods~\citep{wu2025retrieval,zhang2025qrhead,baek2025ocrhead,pei2026vera} mainly differ in their choices of query token set $\mathcal{Q}_s$, key aggregation function $\Phi$, and cross-sample aggregation function $\Omega$.
We next briefly introduce the common design choices for these variables used in prior work~\citep{wu2025retrieval,zhang2025qrhead,pei2026vera,kang2024locheads,baek2025ocrhead}.

\begin{itemize}[leftmargin=*]
\item \textbf{Query token set $\mathcal{Q}_{s}$}: (i)
\textit{Input query tokens} $\mathcal{Q}^{\mathrm{in}}_{s}$, which specify the retrieval target, or (ii) \textit{output tokens} $\mathcal{Q}^{\mathrm{out}}_{s}$, where the retrieved information is generated.

\item \textbf{Key aggregation function $\Phi$}:
(i) \textit{An argmax indicator} $\Phi^{\mathrm{argmax}}$, which returns a binary score for whether the most-attended visible key token falls within $\mathcal{K}_{s}$, or (ii) \textit{a region-sum score} $\Phi^{\mathrm{sum}}$, which returns the total attention mass assigned to key tokens in $\mathcal{K}_{s}$.

\item \textbf{Cross-sample aggregation function $\Omega$}:
(i) \textit{average aggregation} $\Omega^{\mathrm{avg}}$, which averages per-sample scores across the dataset, and (ii) \textit{selection-frequency aggregation} $\Omega^{\mathrm{freq}}$, which ranks heads by how often they are selected across samples.
\end{itemize}

Tab.~\ref{tab:head_scoring_design} summarizes how existing methods instantiate these choices.
Based on this unified view, we define a design space for detecting \Ours{}s: we construct $2^3$ head-scoring variants by varying the query-token set $\mathcal{Q}_s$, key aggregation $\Phi$, and cross-sample aggregation $\Omega$ according to the common choices above.

\vspace{-\baselineskip}
\begin{table}[!h]
\caption{
\textbf{Design choices in head-scoring methods.}
We summarize how existing methods instantiate the main components of Eq.~\ref{eq:head_scoring_template}.
For $\mathcal{Q}_{s}$, some methods use only a few representative tokens from the indicated input or output token set, i.e., $\mathcal{Q}_s^{\texttt{in}}$ or $\mathcal{Q}_s^{\texttt{out}}$; we slightly abuse notation for abstraction.
$^{\dagger}$Localization Heads is a special case that does not directly use key aggregation, but instead selects heads based on the spatial entropy of attention weights.
}
\vspace{0.5\baselineskip}
\centering
\small
\setlength{\tabcolsep}{4pt}
\resizebox{\linewidth}{!}{\begin{tabular}{lccccc}
\toprule
&
\textbf{Retrieval Heads~\citep{wu2025retrieval}}
& \textbf{QRHead~\citep{zhang2025qrhead}}
& \textbf{OCR Heads~\citep{baek2025ocrhead}}
& \textbf{VERA~\citep{pei2026vera}}
& \textbf{Loc. Head~\citep{kang2024locheads}} \\
\midrule

\textbf{Query Token Set $\mathcal{Q}_s$}
& $\mathcal{Q}_s^{\texttt{out}}$
& $\mathcal{Q}_s^{\texttt{in}}$
& $\mathcal{Q}_s^{\texttt{out}}$
& $\mathcal{Q}_s^{\texttt{out}}$
& $\mathcal{Q}_s^{\texttt{in}}$ \\

\textbf{Key Aggregation $\Phi$}
& $\Phi^{\texttt{argmax}}$
& $\Phi^{\texttt{sum}}$
& $\Phi^{\texttt{argmax}}$
& $\Phi^{\texttt{sum}}$
& N/A$^{\dagger}$ \\

\textbf{Sample Aggregation $\Omega$}
& $\Omega^{\texttt{avg}}$
& $\Omega^{\texttt{avg}}$
& $\Omega^{\texttt{avg}}$
& $\Omega^{\texttt{avg}}$
& $\Omega^{\texttt{freq}}$ \\

\bottomrule
\end{tabular}}
\vspace{-0.5\baselineskip}
\label{tab:head_scoring_design}
\end{table}

\subsection{Causal Validation of Head-Scoring Variants}
\label{subsec:causal_validation}

The unified scoring formulation above defines a family of candidate scoring methods for identifying \Ours{}s, depending on the choice of the set of query tokens, key aggregation, and cross-sample aggregation.
While these scores measure how strongly a head's attention aligns with the target visual region, attention alignment alone does not establish that the head is functionally necessary for visual retrieval.
As emphasized in prior work~\cite{zhang2025qrhead,wu2025retrieval,baek2025ocrhead,pei2026vera}, \textit{causality} is essential for interpreting a head as a meaningful functional unit rather than an attention-correlated artifact.
We therefore perform \emph{causal validation}: for each scoring variant, we select the top-ranked heads, mask them during inference, and measure the resulting drop in visual grounding accuracy.

\begin{figure}[!h]
  \centering

  \includegraphics[width=\linewidth]{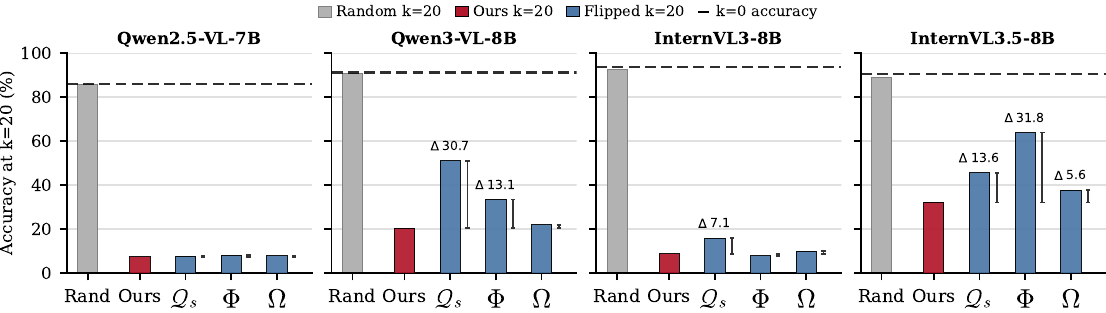}

\caption{
\textbf{Causal validation of head-scoring variants on RefCOCO~\cite{yu2016refcoco} across four VLMs~\cite{bai2025qwen25vl,bai2025qwen3vl,zhu2025internvl3,wang2025internvl35}.}
We mask the top-20 heads detected by each variant and report visual grounding accuracy; lower accuracy indicates that the detected heads are more causally important.
\textit{Ours} denotes our default variant $(\mathcal{Q}^{\mathrm{out}}_s, \Phi^{\mathrm{sum}}, \Omega^{\mathrm{mean}})$.
The \textbf{$\mathcal{Q}_s$}, \textbf{$\Phi$}, and \textbf{$\Omega$} ablations each change one component of this default design at a time, while \textit{Random} masks 20 randomly selected heads.
The dashed line shows the unmasked model ($k=0$), and $\Delta$ denotes the accuracy gap relative to \textit{Ours}.
}
\vspace{-1\baselineskip}
\label{fig:score_ablation}
\end{figure}

We perform this causal validation and empirically choose $(\mathcal{Q}^{\mathrm{out}}_s, \Phi^{\mathrm{sum}}, \Omega^{\mathrm{mean}})$ as the default design for \Ours{} detection.
Fig.~\ref{fig:score_ablation} reports this validation across four VLMs~\cite{bai2025qwen25vl,bai2025qwen3vl,zhu2025internvl3,wang2025internvl35} on RefCOCO~\cite{yu2016refcoco}, ablating one design component at a time and measuring whether the resulting top-ranked heads are causally important for visual grounding.
Masking heads detected by the default variant consistently causes large accuracy drops relative to the unmasked model ($k=0$), showing that this score identifies heads that are causally important for visual grounding.
The ablations further show that the query-token set is the most important design choice:
variants that score attention from output query tokens identify heads whose removal causes substantially larger and more consistent accuracy drops than variants based on input query tokens.
This suggests that the strongest causal retrieval signal emerges when the model is autoregressively producing the grounded answer, rather than when it only encodes the referring expression in the prompt.

\paragraph{Comparison with existing specialized heads.}

\begin{figure}[!h]
  \centering

  \includegraphics[width=\linewidth]{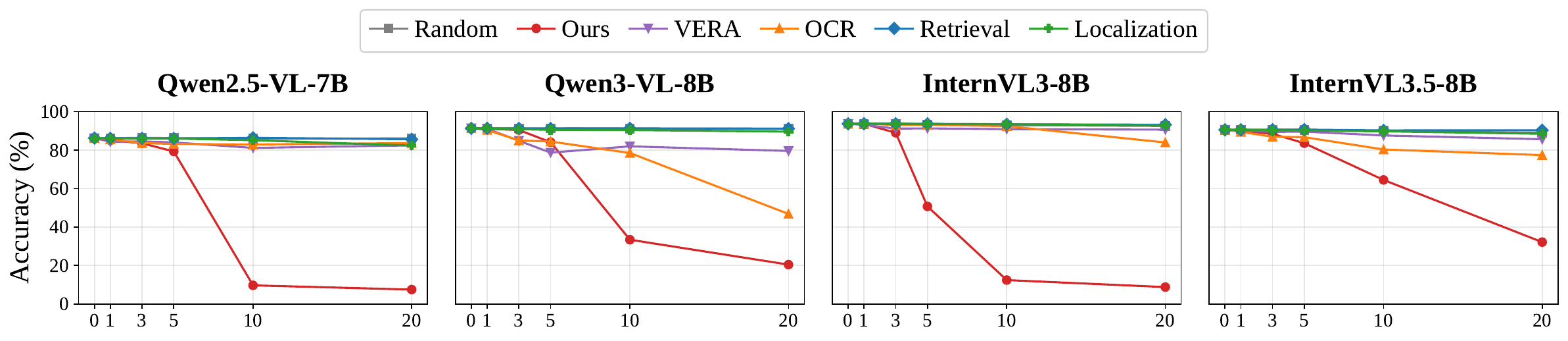}

\caption{
\textbf{Comparison with existing heads.}
We compare VRHs with Retrieval Heads~\citep{wu2025retrieval}, OCR Heads~\citep{baek2025ocrhead}, VERA~\citep{pei2026vera}, Localization Heads~\citep{kang2024locheads}, and randomly selected heads on the RefCOCO visual grounding benchmark~\citep{yu2016refcoco}.
The x-axis denotes the number of masked heads ($k$). Masking the top-$k$ VRHs causes substantially larger drops in grounding accuracy than masking these baselines, indicating that VRHs capture a distinct set of heads causally important for visual grounding.
}
\label{fig:head_comparison}
\vspace{-1.2\baselineskip}
\end{figure}

We next ask whether \Ours{}s simply overlap with specialized heads identified by prior methods, or instead reveal a distinct mechanism for visual reference resolution in general images.
As discussed in Sec.~\ref{sec:related}, most existing methods focus on text retrieval~\cite{wu2025retrieval,zhang2025qrhead} or visual retrieval from document-like images~\cite{baek2025ocrhead,pei2026vera}.
Localization Heads~\citep{kang2024locheads} move beyond rendered text toward general images, but characterize heads by spatial concentration rather than by their functional role.
In contrast, \Ours{}s are identified by directly scoring how strongly each head attends from output tokens to the target image region, and are validated by their \emph{causal} effect on visual retrieval performance.
Using the same causal validation protocol on RefCOCO~\citep{yu2016refcoco}, Fig.~\ref{fig:head_comparison} shows that masking \Ours{}s causes the largest degradation in grounding accuracy across four VLMs, substantially exceeding the effect of masking random heads or those identified by prior methods.

\vspace{-0.2\baselineskip}

\section{Visual Retrieval Heads Are Universal and Sparse}
\label{sec:empirical}

Sec.~\ref{subsec:causal_validation} establishes that our scoring method identifies heads that are causally important for visual grounding.
We now ask a stronger question: are these heads merely an artifact of a particular model, dataset, or scoring setup, or do they reveal a general mechanism by which VLMs perform visual retrieval? Following the criteria used to characterize retrieval heads in LLMs~\citep{wu2025retrieval,zhang2025qrhead}, we evaluate whether \Ours{}s are both \emph{universal} and \emph{sparse}.

\begin{figure}[!h]
  \centering

  \newcommand{\rowlabelfont}{\footnotesize}

  \newlength{\plotwidth}
  \setlength{\plotwidth}{0.95\linewidth}

  \setlength{\tabcolsep}{2pt}
  \renewcommand{\arraystretch}{0}

  \begin{tabular}{@{}c@{\hspace{2pt}}c@{}}
    \raisebox{0.2\height}{\rotatebox{90}{\rowlabelfont RefCOCO (avg)}} &
    \includegraphics[width=\plotwidth]{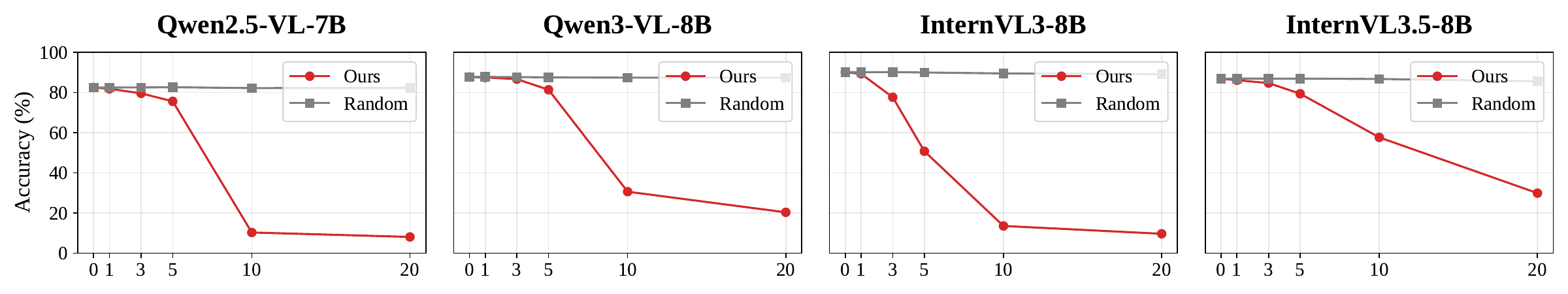} \\
    \raisebox{0.4\height}{\rotatebox{90}{\rowlabelfont RefSpatial}} &
    \includegraphics[width=\plotwidth]{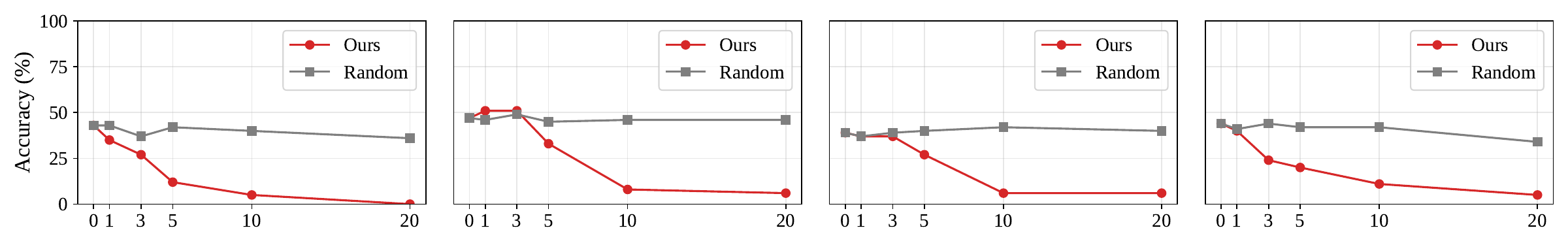} \\
    \raisebox{0.8\height}{\rotatebox{90}{\rowlabelfont Toloka}} &
    \includegraphics[width=\plotwidth]{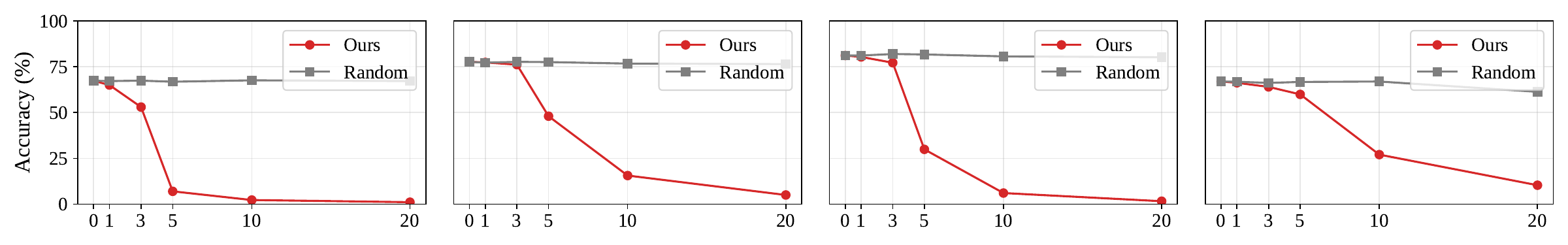} \\
  \end{tabular}

\caption{
\textbf{Universality of VRHs across models and benchmarks.}
We report visual grounding accuracy after masking the top-$k$ VRHs across four VLMs and three benchmark families: RefCOCO/+/g~\cite{yu2016refcoco,mao2016refcocog}, RefSpatialBench~\cite{zhou2025roborefer}, and Toloka~\cite{ustalov2023toloka}.
The x-axis denotes the number of masked heads ($k$). Masking only a few VRHs consistently causes large accuracy drops, while masking random heads has little effect.
This broad causal effect shows that VRHs are not architecture- or dataset-specific, but instead capture a shared attention-head mechanism for visual grounding.
}
  \label{fig:universality}
  \vspace{-\baselineskip}
\end{figure}

\paragraph{Universality.}
\Ours{}s are universal, consistently emerging as a causal visual-retrieval unit across diverse model families and grounding benchmarks.
We evaluate eleven VLMs (Fig.~\ref{fig:allmodels}) built on eight language-model backbones (Qwen2.5, Qwen3, Llama-3, Gemma~2, DeepSeekMoE, InternLM2.5, GLM and Phi~\citep{bai2025qwen25vl,bai2025qwen3vl,zhu2025internvl3,wang2025internvl35,yang2025magma,li2024llavanext,steiner2024paligemma2,wu2024deepseekvl2,chen2024internvl25,glm2025glm4v,abouelenin2025phi4rv}) on five grounding datasets spanning three benchmark families: (RefCOCO/+/g~\citep{yu2016refcoco,mao2016refcocog}, RefSpatialBench~\citep{zhou2025roborefer}, and Toloka~\citep{ustalov2023toloka}).
As shown in Fig.~\ref{fig:universality}, masking the top-ranked \Ours{}s consistently causes large drops in grounding accuracy across all model--benchmark pairs, whereas masking the same number of random heads has little effect.
Fig.~\ref{fig:allmodels} summarizes the same contrast at $k{=}20$ on RefCOCO for every evaluated VLM, where the seven models beyond the four studied in depth are evaluated on a separate detection and evaluation split of the benchmark (Sec.~\ref{app:models}); the same detection and masking procedure collapses grounding in all of them.
This indicates that, despite differences in architecture, training recipe, and benchmark distribution, diverse VLMs rely on a sparse set of attention heads for correct visual reference behavior.

\paragraph{Sparsity.}

\begin{wrapfigure}{r}{0.44\linewidth}
  \centering
  \vspace{-0.4em}
  \includegraphics[width=\linewidth]{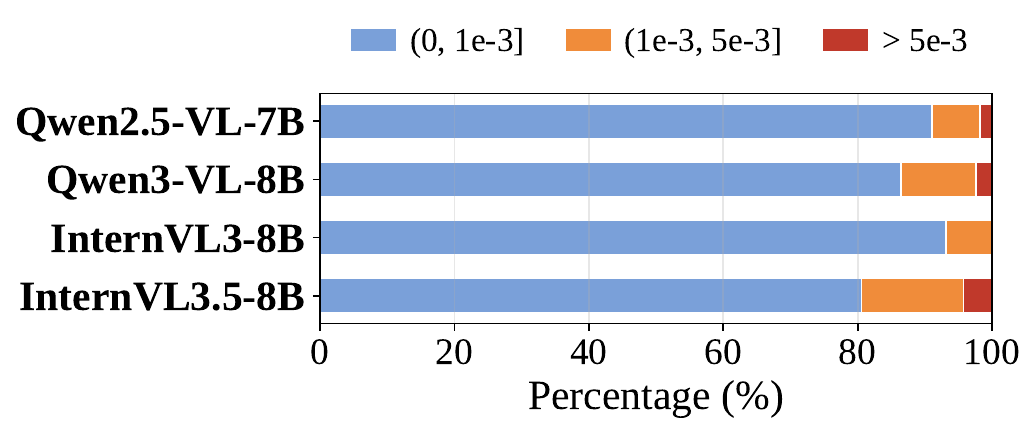}
  \caption{
\textbf{Sparsity of \Ours{}s.}
Distribution of head scores for \Ours{} detection on RefCOCO~\cite{yu2016refcoco} across four VLMs~\cite{bai2025qwen25vl,bai2025qwen3vl,zhu2025internvl3,wang2025internvl35}. Only a small fraction of heads receive non-negligible scores.
}
  \label{fig:sparsity_barplot}
  \vspace{-1.0em}
\end{wrapfigure}

\Ours{}s are highly sparse, suggesting that visual retrieval is concentrated in a small set of specialized heads rather than diffusely implemented throughout the transformer.
Fig.~\ref{fig:sparsity_barplot} shows the distribution of \Ours{} scores on RefCOCO across the four VLMs.
Fewer than 15\% of heads receive a non-negligible score (above $10^{-3}$), and only 0--4\% receive a high score (above $5{\times}10^{-3}$), depending on the model.
Despite occupying only a small fraction of each model, these heads have a large causal effect: masking them is sufficient to substantially disrupt grounding.
Thus, visual retrieval in VLMs mirrors the sparse-but-causal organization of retrieval heads in LLMs~\citep{wu2025retrieval}.

\begin{figure}[!h]
  \centering
  \includegraphics[width=\linewidth]{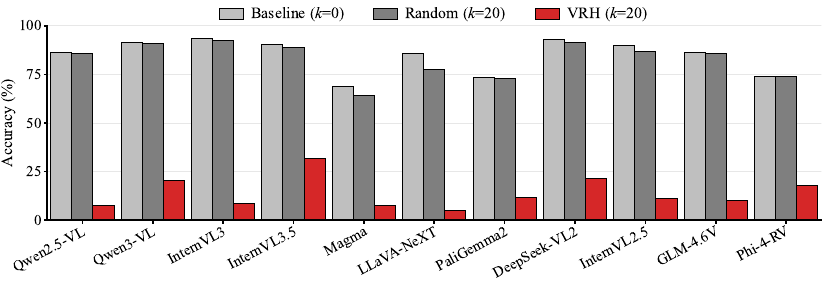}
\caption{
\textbf{\Ours{}s are universal across VLMs.}
Visual grounding accuracy on RefCOCO~\cite{yu2016refcoco} for all eleven evaluated VLMs, before masking, after masking 20 random heads, and after masking the top-20 \Ours{}s. Masking the \Ours{}s causes a large accuracy drop in every model, whereas masking the same number of random heads has little effect. Exact values, evaluation splits and per-model protocols are given in Sec.~\ref{app:models}.
}
  \label{fig:allmodels}
  \vspace{-0.5\baselineskip}
\end{figure}

\vspace{-0.4\baselineskip}

\section{Characterizing Visual Retrieval Heads}
\label{sec:characterization}

In previous sections (Sec.~\ref{sec:method}--\ref{sec:empirical}), we established that \Ours{}s are causal, universal, and sparse---properties shared with existing retrieval heads for text or visual text~\citep{wu2025retrieval,baek2025ocrhead,pei2026vera,zhang2025qrhead}.
We now show that \Ours{}s exhibit important additional properties:
(i)~\Ours{}s \emph{generalize across visual reference tasks} beyond the grounding task used to detect them (Sec.~\ref{sec:char_generalization});
(ii)~\Ours{}s are \emph{functionally specific} to visual retrieval, selectively disrupting reference resolution without degrading general generation (Sec.~\ref{sec:functional_specificity});
(iii)~\Ours{}s are \emph{architecturally shared} across VLMs built on the same LLM backbone, indicating they reside in a reusable language-model circuit (Sec.~\ref{sec:architecturally_shared});
and (iv)~\Ours{}s are \emph{stably detectable}, remaining consistent across output-token choices and requiring only a handful of annotated examples (Sec.~\ref{sec:stably_detectable}).
Note that, while we mainly focus on characterizing the properties of \Ours{}s in this section, we kindly refer readers to Sec. B in the Appendix, where we also show that \Ours{}s can also be utilized to improve visual grounding accuracy via direct preference optimization (DPO).

\vspace{-0.2\baselineskip}

\subsection{VRHs Generalize Across Visual Reference Tasks}
\label{sec:char_generalization}

We first show that \Ours{}s effectively generalize beyond visual grounding: they implement a more general mechanism for visual reference resolution required by diverse vision question-answering tasks involving attributes, spatial reasoning, counting, vision-centric understanding, and visual math.
This cross-task generalization importantly distinguishes \Ours{}s from specialized heads studied in prior work on LLMs and VLMs~\citep{wu2025retrieval,zhang2025qrhead,baek2025ocrhead,pei2026vera}, which are typically identified and evaluated within a single task, leaving open whether the discovered heads support a broader cognitive function or only the particular evaluation format used to find them.

We validate this by taking \Ours{}s detected on a visual grounding benchmark, RefCOCO~\citep{yu2016refcoco}, and testing their causal effects on seven VQA benchmarks spanning the aforementioned tasks: VAW~\citep{pham2021vaw}, Spatial457~\citep{wang2025spatial457}, SpatialRGPT~\citep{cheng2024spatialrgpt}, CV-Bench~\citep{tong2024cvbench}, MMStar~\citep{chen2024mmstar}, CountBenchQA~\citep{paiss2023countbench}, and MathVista~\citep{lu2023mathvista}.
As shown in Tab.~\ref{tab:downstream}, masking \Ours{}s consistently degrades performance across all seven benchmarks and all four VLMs, while masking the same number of random heads has little effect.
These results reveal that, although \Ours{}s are discovered from visual grounding prompts that require generating \textit{bounding-box coordinates} for the referred object, they capture a general visual reference mechanism required across diverse tasks.

\definecolor{oursred}{RGB}{180,40,40}
\definecolor{randgray}{RGB}{95,95,95}

\newcommand{\drop}[1]{{\scriptsize\textcolor{oursred}{(#1)}}}
\newcommand{\rdrop}[1]{{\scriptsize\textcolor{randgray}{(#1)}}}

\begin{table}[!h]
\centering
\caption{
\textbf{Generalization across visual reference tasks.}
Masking the top-20 VRHs detected on RefCOCO~\cite{yu2016refcoco} consistently degrades performance on seven VQA benchmarks~\cite{pham2021vaw, wang2025spatial457, cheng2024spatialrgpt, tong2024cvbench, chen2024mmstar, paiss2023countbench, lu2023mathvista} spanning attributes,
spatial reasoning, counting, vision-centric understanding, and visual math, while random-head masking has little effect.
}
\label{tab:downstream}
\vspace{0.5\baselineskip}
\resizebox{\columnwidth}{!}{\begin{tabular}{ll*{7}{l}}
\toprule
Model & Condition & VAW & Spatial457 & SpatialRGPT & CV-Bench & MMStar & CountBenchQA & MathVista \\
\midrule
\multirow{3}{*}{Qwen2.5-VL-7B}
 & Baseline ($k{=}0$)
 & 63.8
 & 68.2
 & 36.1
 & 75.1
 & 58.3
 & 83.9
 & 71.5 \\
 & Random ($k{=}20$)
 & 63.7 \rdrop{-0.1}
 & 67.2 \rdrop{-1.0}
 & 39.5 \rdrop{+3.4}
 & 71.9 \rdrop{-3.2}
 & 56.8 \rdrop{-1.5}
 & 84.1 \rdrop{+0.2}
 & 72.0 \rdrop{+0.5} \\
 & \Ours{} ($k{=}20$)
 & 40.5 \drop{-23.3}
 & 36.0 \drop{-32.2}
 & 21.8 \drop{-14.3}
 & 53.3 \drop{-21.8}
 & 47.9 \drop{-10.4}
 & 61.5 \drop{-22.4}
 & 61.7 \drop{-9.8} \\
\midrule
\multirow{3}{*}{Qwen3-VL-8B}
 & Baseline ($k{=}0$)
 & 74.8
 & 78.2
 & 50.8
 & 87.6
 & 63.0
 & 88.4
 & 66.5 \\
 & Random ($k{=}20$)
 & 74.3 \rdrop{-0.5}
 & 76.4 \rdrop{-1.8}
 & 46.2 \rdrop{-4.6}
 & 87.1 \rdrop{-0.5}
 & 63.3 \rdrop{+0.3}
 & 89.6 \rdrop{+1.2}
 & 69.4 \rdrop{+2.9} \\
 & \Ours{} ($k{=}20$)
 & 51.0 \drop{-23.8}
 & 36.9 \drop{-41.3}
 & 22.9 \drop{-27.9}
 & 68.7 \drop{-18.9}
 & 47.7 \drop{-15.3}
 & 74.5 \drop{-13.9}
 & 43.0 \drop{-23.5} \\
\midrule
\multirow{3}{*}{InternVL3-8B}
 & Baseline ($k{=}0$)
 & 73.1
 & 73.1
 & 50.0
 & 85.9
 & 67.9
 & 81.3
 & 80.0 \\
 & Random ($k{=}20$)
 & 71.9 \rdrop{-1.2}
 & 72.4 \rdrop{-0.7}
 & 49.6 \rdrop{-0.4}
 & 86.7 \rdrop{+0.8}
 & 67.7 \rdrop{-0.2}
 & 81.9 \rdrop{+0.6}
 & 79.8 \rdrop{-0.2} \\
 & \Ours{} ($k{=}20$)
 & 47.6 \drop{-25.5}
 & 37.3 \drop{-35.8}
 & 38.0 \drop{-12.0}
 & 69.3 \drop{-16.6}
 & 47.6 \drop{-20.3}
 & 60.1 \drop{-21.2}
 & 65.7 \drop{-14.3} \\
\midrule
\multirow{3}{*}{InternVL3.5-8B}
 & Baseline ($k{=}0$)
 & 72.6
 & 66.3
 & 50.4
 & 80.3
 & 64.9
 & 78.0
 & 78.9 \\
 & Random ($k{=}20$)
 & 72.0 \rdrop{-0.6}
 & 64.3 \rdrop{-2.0}
 & 51.5 \rdrop{+1.1}
 & 82.8 \rdrop{+2.5}
 & 64.4 \rdrop{-0.5}
 & 79.0 \rdrop{+1.0}
 & 79.4 \rdrop{+0.5} \\
 & \Ours{} ($k{=}20$)
 & 54.4 \drop{-18.2}
 & 39.3 \drop{-27.0}
 & 41.0 \drop{-9.4}
 & 68.6 \drop{-11.7}
 & 51.5 \drop{-13.4}
 & 65.2 \drop{-12.8}
 & 66.3 \drop{-12.6} \\
\bottomrule
\end{tabular}
}
\vspace{-\baselineskip}
\end{table}

\vspace{-0.3\baselineskip}

\subsection{VRHs are Functionally Specific}
\label{sec:functional_specificity}

We next show that \Ours{}s are functionally specific: masking them does not simply degrade generation, but selectively disrupts visual reference resolution.
For example, on the visual grounding task, we observe that masking \Ours{}s leads to mislocalized bounding-box predictions, while the bounding-box syntax remains correct.
In Fig.~\ref{fig:failure_case}, we further decompose visual-grounding failures under \Ours{} masking versus random-head masking into two types: (a) \emph{parse failure}, where the output does not preserve the bounding-box syntax and cannot be parsed, and (b) \emph{mislocation}, where the output is a syntactically valid bounding box but is spatially mislocalized.

We find that masking \Ours{}s significantly increases mislocation rates while parse failures remain rare. 
This contrast mirrors the ``fluent-but-not-factual'' failure mode reported for text retrieval heads~\citep{wu2025retrieval,zhang2025qrhead}: masking the specialized heads preserves the surface form of the output, but corrupts the factuality of the evidence-grounded content. Fig.~\ref{fig:teaser} illustrates this failure mode qualitatively: masked models produce syntactically valid bounding boxes that localize incorrect regions.

\subsection{VRHs are Architecturally Shared}
\label{sec:architecturally_shared}

\begin{figure}[!h]
  \centering
  \includegraphics[width=\linewidth]{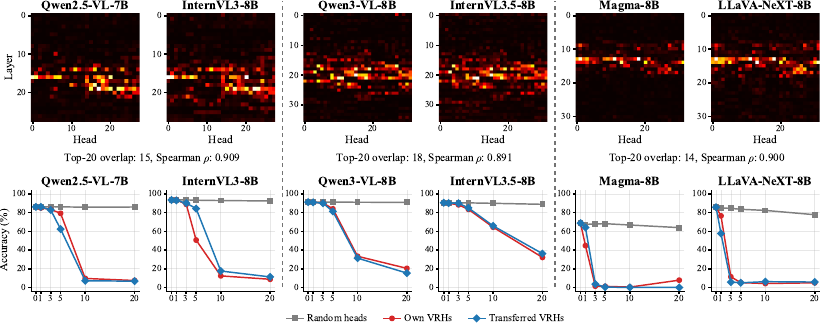}

\caption{
\textbf{\Ours{} detection consistency across same-backbone VLMs.}
Each block is a pair of VLMs that share a language backbone but differ in vision encoder, projector and instruction tuning: Qwen2.5-VL-7B~\cite{bai2025qwen25vl} with InternVL3-8B~\cite{zhu2025internvl3}, which share a Qwen2.5-7B backbone; Qwen3-VL-8B~\cite{bai2025qwen3vl} with InternVL3.5-8B~\cite{wang2025internvl35}, which share a Qwen3-8B backbone; and Magma-8B~\cite{yang2025magma} with LLaVA-NeXT-Llama3-8B~\cite{li2024llavanext}, which share a Llama-3 backbone.
\textbf{Top row:} Layer-head score maps, where each cell denotes the VRH score $\bar{\Theta}_{l,h}$ at layer $l$ and head $h$; high-scoring heads appear at similar layer-head locations across same-backbone VLMs.
\textbf{Bottom row:} Cross-model masking results, where the top-k VRHs identified in one model are masked at the same layer-head indices in the other model. \emph{Own} \Ours{}s are the heads detected in the panel's own model and \emph{transferred} \Ours{}s those detected in its partner. The x-axis denotes the number of masked heads (k).
}
\label{fig:same_backbone_consistency}
\end{figure}

\begin{figure}[b]
  \centering
  \begin{subfigure}[t]{0.34\linewidth}
    \centering
    \includegraphics[width=\linewidth]{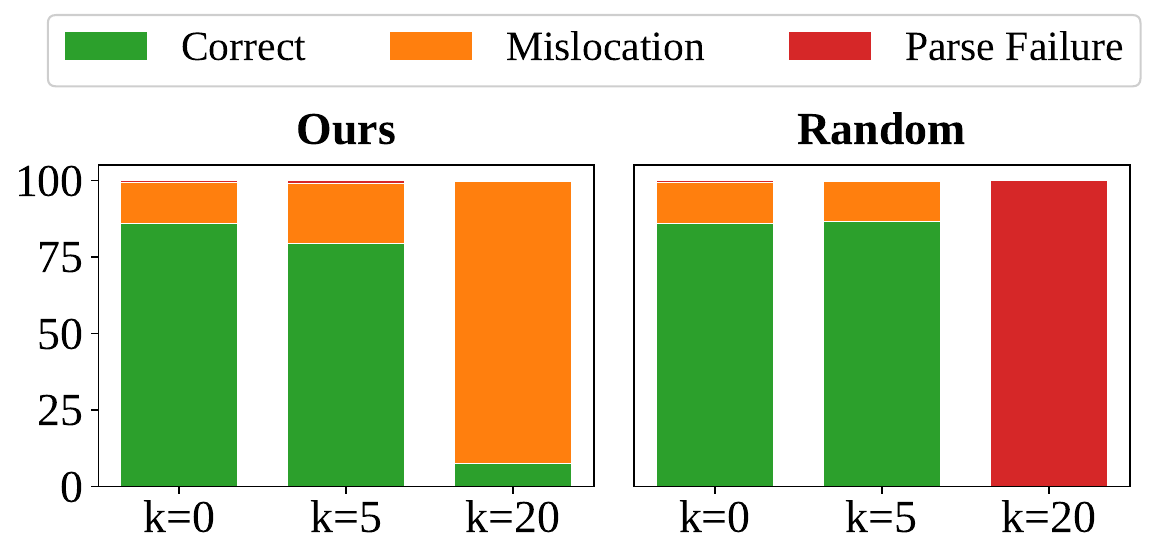}
    \caption{Functional specificity}
    \label{fig:failure_case}
  \end{subfigure}
  \hfill
  \begin{subfigure}[t]{0.3\linewidth}
    \centering
    \includegraphics[width=\linewidth]{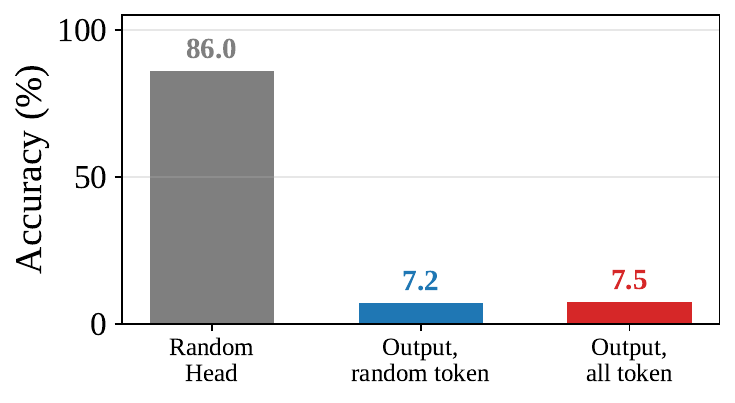}
    \caption{Output-token robustness}
    \label{fig:token_jaccard}
  \end{subfigure}
  \hfill
  \begin{subfigure}[t]{0.32\linewidth}
    \centering
    \includegraphics[width=\linewidth]{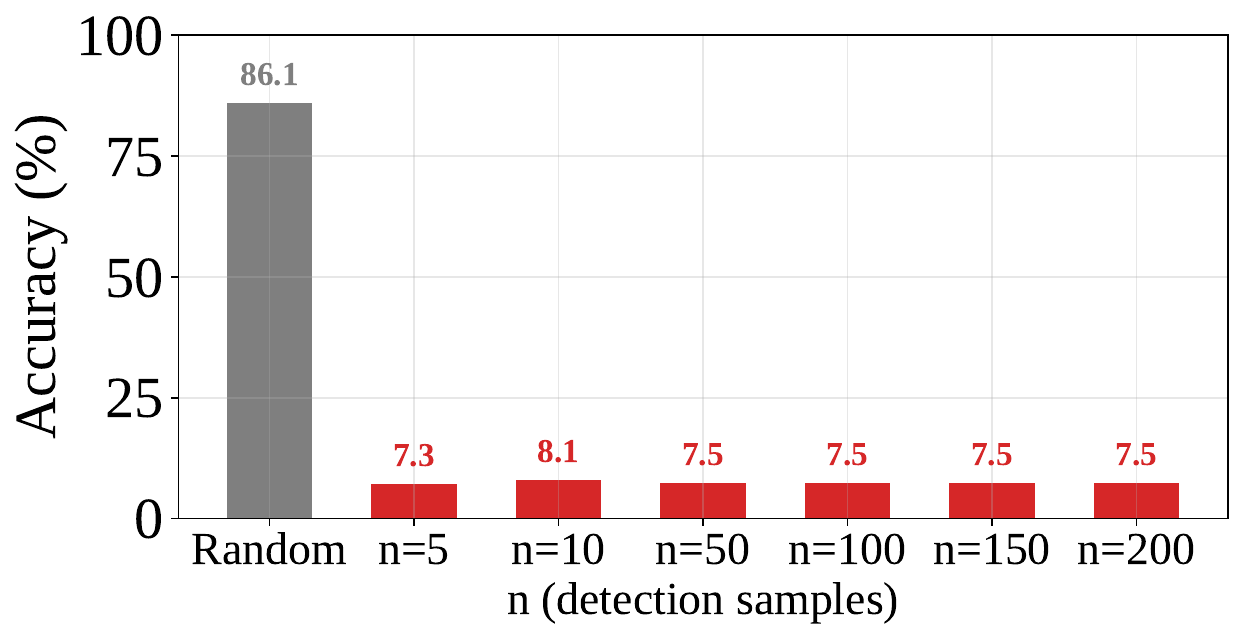}
    \caption{Sample-size robustness}
    \label{fig:sample_efficiency}
  \end{subfigure}
\caption{
\textbf{VRHs are functionally specific and stably detectable.}
All experiments use Qwen2.5-VL-7B~\cite{bai2025qwen25vl} on RefCOCO~\cite{yu2016refcoco} with $k{=}20$ masked heads unless otherwise noted.
\textbf{(a)}~Visual grounding failure decomposition under VRH masking vs.\ random-head masking. \textit{Mislocation} denotes syntactically valid but mislocalized box prediction, while \textit{parse failure} denotes output that cannot be parsed as a bounding box.
\textbf{(b)}~Grounding accuracy when masking VRHs detected from a single random output token, all output tokens, or random heads.
\textbf{(c)}~Grounding accuracy when masking VRHs detected with varying numbers of scoring samples $n$.}
  \label{fig:specificity_stability}
\end{figure}

We next show that VRHs are not specific to a single VLM instantiation, but reflect an architecturally shared mechanism within the LLM backbone.
In particular, VRHs detected in one VLM remain meaningful in another VLM that shares the same LLM backbone, even when the vision encoder, projector, and instruction tuning differ.
We test this with pairs of same-backbone VLMs in two ways.
First, we compare the \emph{layer-head score map}, where each entry at layer $l$ and head $h$ records the dataset-level \Ours{} score $\bar{\Theta}_{l,h}$.
As shown in Fig.~\ref{fig:same_backbone_consistency}(a), same-backbone VLMs exhibit similar high-scoring layer-head locations, with substantial top-20 overlap and high rank correlation.
Second, we perform \emph{cross-model masking}: we detect the top-20 \Ours{}s in one VLM and mask the same layer-head indices in another.
As shown in Fig.~\ref{fig:same_backbone_consistency}(b), transferred \Ours{}s cause a large grounding-accuracy drop in the target model, approaching the effect of masking its own \Ours{}s and far exceeding random-head masking. The same holds for the third pair in Fig.~\ref{fig:same_backbone_consistency}, Magma-8B~\citep{yang2025magma} and LLaVA-NeXT-Llama3-8B~\citep{li2024llavanext}, where cross-masking the top-20 \Ours{}s reduces grounding accuracy from 68.8\% to 0.0\% and from 85.8\% to 5.8\% respectively.
In summary, these results show that \Ours{}s are architecturally shared across same-backbone VLMs, suggesting that visual retrieval relies on a reusable circuit in the language backbone rather than an idiosyncratic VLM-specific implementation.

\vspace{-0.8\baselineskip}

\subsection{VRHs are Stably Detectable}
\label{sec:stably_detectable}
We finally show that \Ours{}s can be detected reliably across two key sources of variation in the scoring variables: the choice of output query tokens and the number of annotated grounding examples.

\paragraph{Robustness across output tokens.}

\Ours{} detection is stable across output-token query choices: heads identified using different subsets of output tokens are equally causal.
To test this, we select the top-20 \Ours{}s under different query-token choices---a single random output token or all output tokens---and measure grounding accuracy after masking them.
As shown in Fig.~\ref{fig:token_jaccard}, both choices cause a comparable accuracy drop (7.2\% and 7.5\%, respectively), while masking the same number of random heads barely affects performance (86.0\%).
This confirms that our head-scoring method consistently identifies causally important \Ours{}s regardless of the specific output-token query set.

\paragraph{Robustness across sample size.}
\Ours{} detection is highly sample-efficient: only a few grounding annotations are needed to recover the same top-ranked heads as a much larger scoring set.
This indicates that \Ours{}s exhibit a strong, repeatable attention signature, rather than a weak pattern that emerges only after extensive averaging.
To quantify this, we vary the number of samples used to estimate $\bar{\Theta}_{l,h}$ and compare each resulting top-20 head set with the reference set computed from 200 examples.
As shown in Fig.~\ref{fig:sample_efficiency}, as few as 5--10 examples recover nearly the same top-ranked \Ours{}s as the full scoring set.
This makes \Ours{} detection practical: a small number of bounding-box annotations is sufficient to reliably identify \Ours{}s.

\vspace{-0.2\baselineskip}

\FloatBarrier

\vspace{-\baselineskip}
\section{Conclusion}

\vspace{-0.6\baselineskip}

We introduce Visual Retrieval Heads (VRHs), a sparse set of attention heads that are causally necessary for correct visual reference behavior.
Using visual grounding as a probing task, we identify VRHs by scoring attention from output tokens to ground-truth visual regions, and validate their causal role through targeted masking across multiple VLMs and grounding benchmarks.
These analyses show that VRHs are sparse, causal, and universal.
Importantly, their role extends beyond grounding: the same heads are causally important for attribute recognition, spatial reasoning, counting, and visual math, suggesting that VRHs support visual reference resolution generally.

\vspace{-0.5\baselineskip}

\paragraph{Limitation and Future Work.}
\textbf{(i)} Our analysis identifies VRHs through attention weights, following prior retrieval-head methods.
While sufficient to reveal sparse, causal, and general visual-reference heads, this signal primarily captures \emph{where} visual evidence is routed from, rather than \emph{what} information is read out or transformed.
Future work could combine VRH detection with activation patching~\citep{goldowskydill2023pathpatching}, or causal mediation~\citep{li2026causaltracing} for a finer-grained mechanistic account.
\textbf{(ii)} Our experiments focus on single-image VLM inputs as a controlled setting for isolating visual reference behavior.
Extending VRH analysis to multi-image, video, and interleaved image-text contexts could further reveal how VLMs route visual evidence across richer multimodal inputs, with potential applications in efficient visual-context selection~\cite{wang2025cdviews} and multimodal cache compression~\cite{ge2024model}.

\appendix
\newpage
\appendix

\renewcommand{\thefigure}{A\arabic{figure}}
\renewcommand{\thetable}{A\arabic{table}}
\renewcommand{\theequation}{A\arabic{equation}}
\setcounter{figure}{0}
\setcounter{table}{0}
\setcounter{equation}{0}
\setcounter{section}{0}

\section{Additional Qualitative Results}
\label{app:qualitative}

\subsection{Visual Grounding}
\begin{figure}[!htbp]
  \centering

  \begin{tabular}{@{}c@{}}
    \makebox[\linewidth]{%
      \makebox[0.25\linewidth]{Input}%
      \makebox[0.25\linewidth]{All heads}%
      \makebox[0.25\linewidth]{VRHs}%
      \makebox[0.25\linewidth]{All heads except VRHs}%
    } \\
    \midrule
    \includegraphics[width=\linewidth]{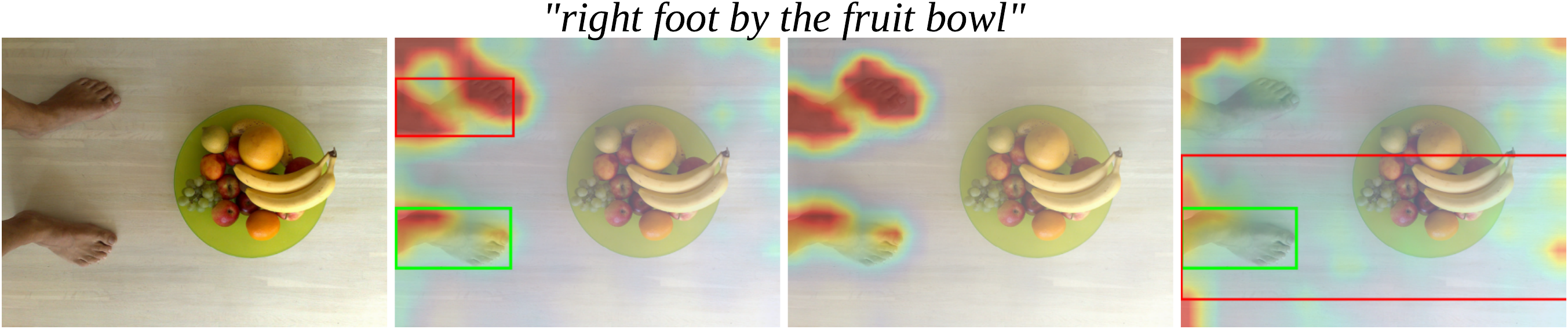} \\
    \includegraphics[width=\linewidth]{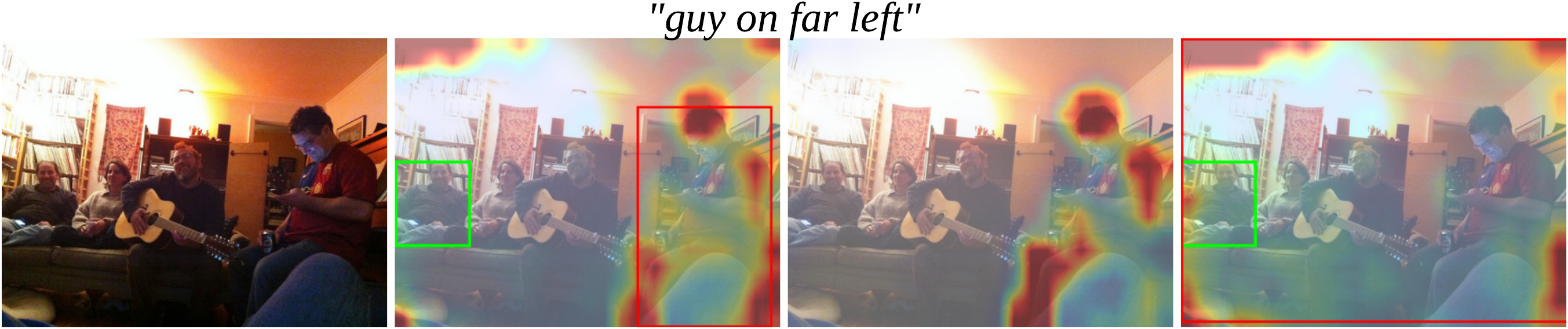} \\
    \includegraphics[width=\linewidth]{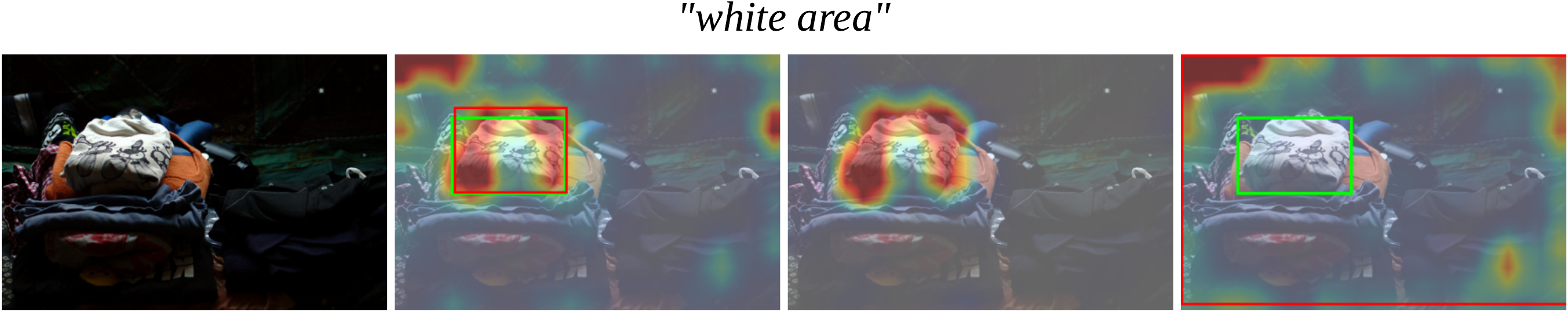} \\
    \includegraphics[width=\linewidth]{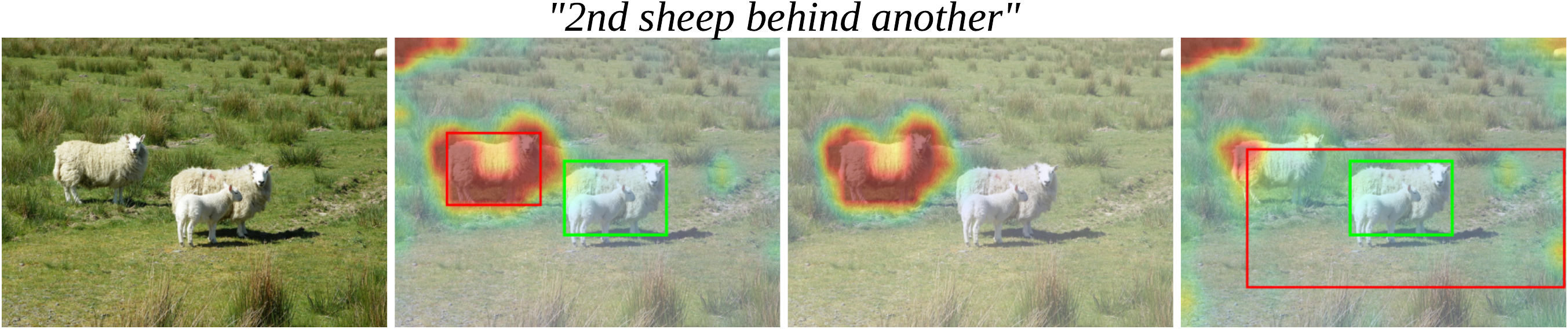} \\
    \includegraphics[width=\linewidth]{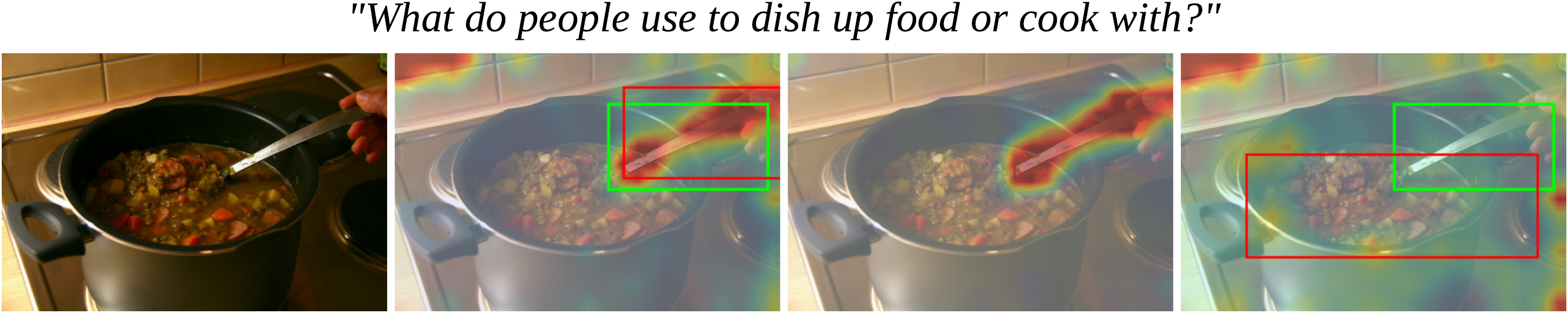} \\
  \end{tabular}

\caption{
\textbf{Attention map visualizations on visual grounding tasks.}
From left to right, the columns show the input image and the mean attention maps over: (i) all heads, (ii) the top-20 \Ours{}s, and (iii) all heads excluding \Ours{}s.
Green and red boxes indicate ground-truth and predicted bounding boxes, respectively.
}
\label{fig:quali_grounding}
\end{figure}

In Figure~\ref{fig:quali_grounding}, we provide additional visualizations that complement Fig.~1 in the main paper.
Specifically, we visualize the average attention map under three attention-head settings: (i) all heads, (ii) \Ours{}s, and (iii) all heads excluding \Ours{}s, on two grounding benchmarks: RefCOCO~\citep{yu2016refcoco} and Toloka~\citep{ustalov2023toloka}. 
Across all examples, the all-heads attention concentrates on the ground-truth referent (column ii), and this concentration is preserved when only \Ours{}s are kept (column iii) but largely lost when \Ours{}s are masked (column iv).
\Ours{}s are therefore the heads responsible for routing attention to the referent: under \Ours{} masking, attention on the referent vanishes and the predicted bounding box becomes mislocalized.

\FloatBarrier

\newpage
\subsection{Visual Question Answering}
\begin{figure}[!htbp]
  \centering

  \begin{minipage}{0.49\linewidth}
    \centering
    \includegraphics[width=\linewidth]{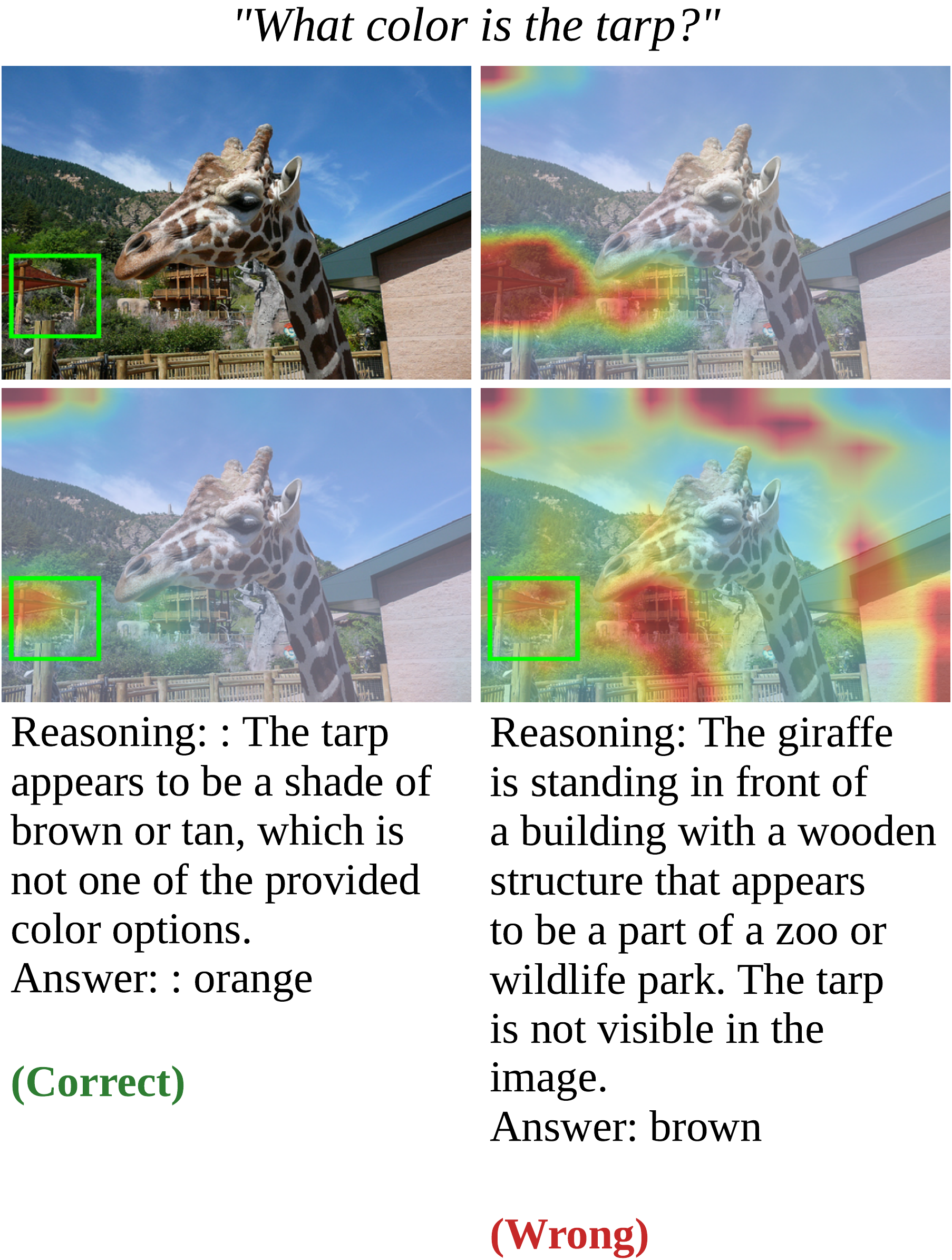}
  \end{minipage}
  \hfill
  \begin{minipage}{0.49\linewidth}
    \centering
    \includegraphics[width=\linewidth]{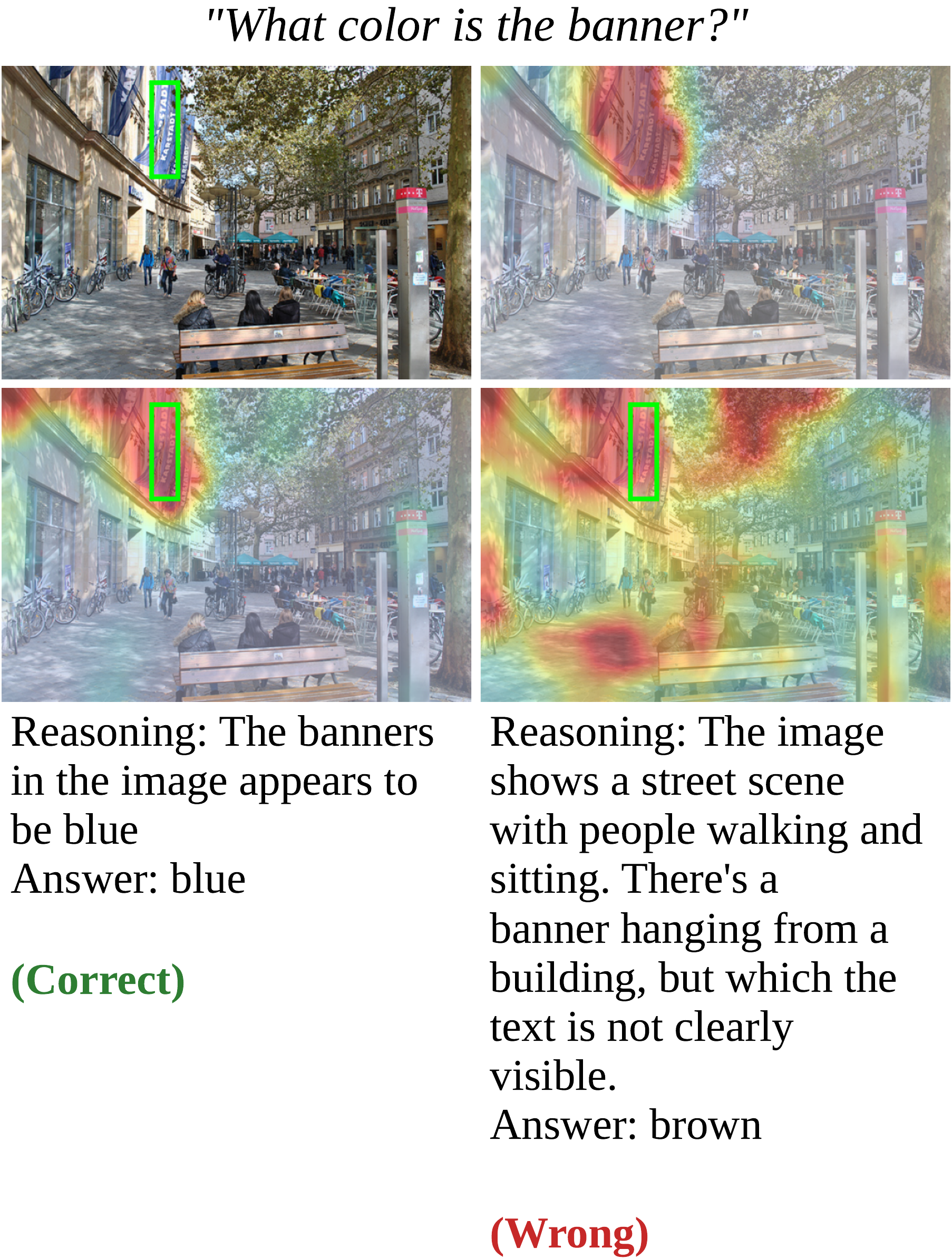}
  \end{minipage}

\vspace{1em}

  \begin{minipage}{0.49\linewidth}
    \centering
    \includegraphics[width=\linewidth]{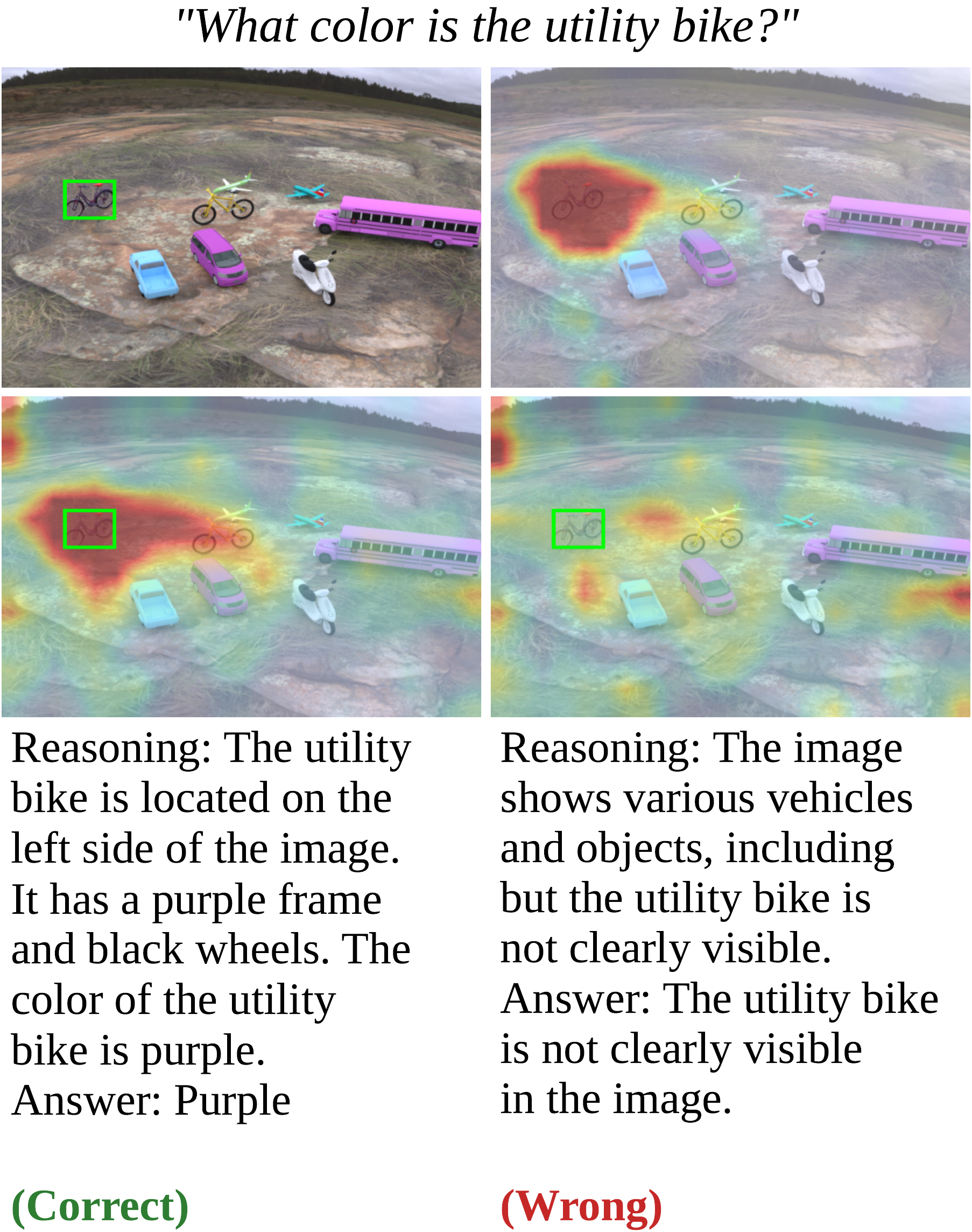}
  \end{minipage}
  \hfill
  \begin{minipage}{0.49\linewidth}
    \centering
    \includegraphics[width=\linewidth]{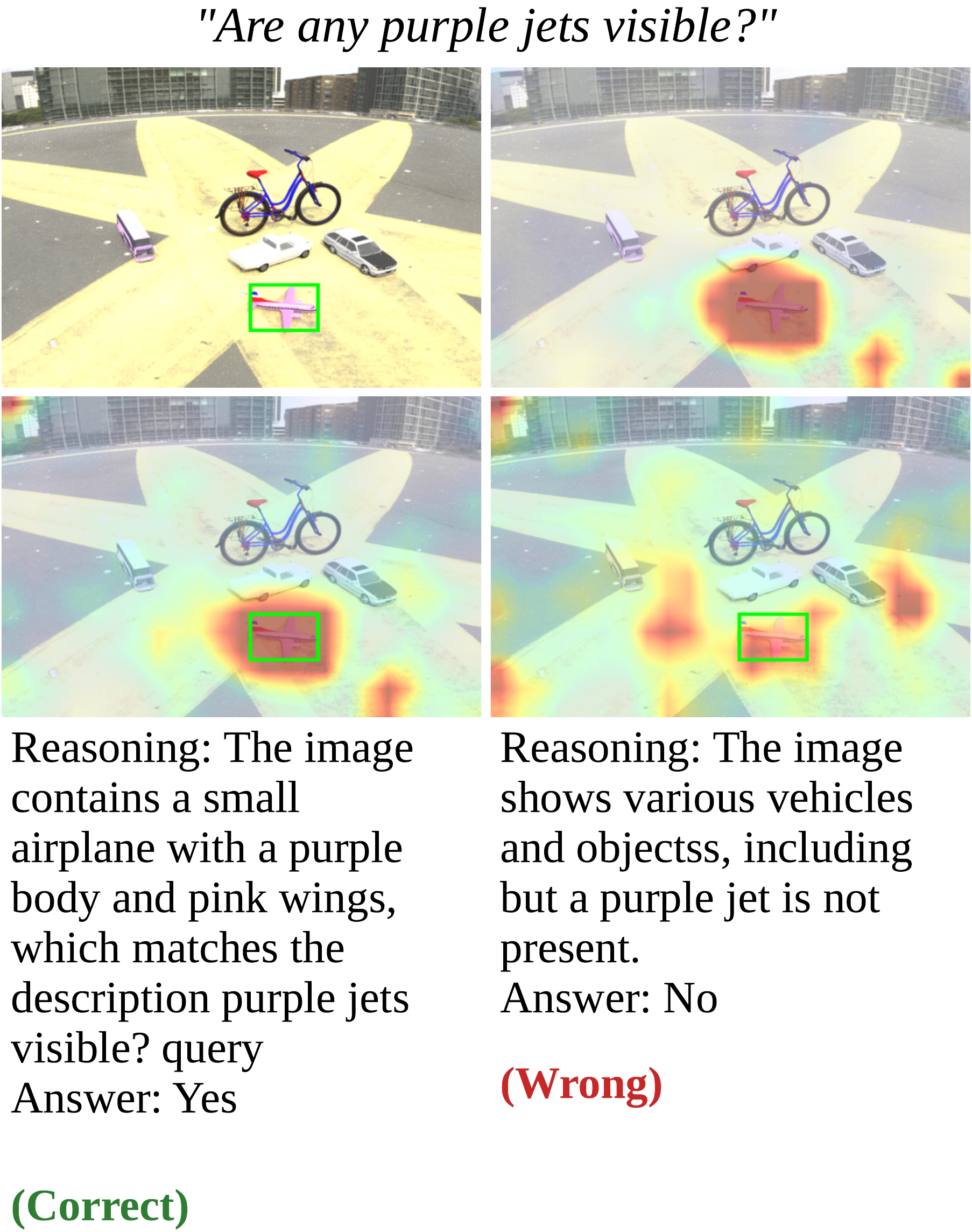}
  \end{minipage}

  \caption{
  \textbf{Attention map visualizations on visual question-answering tasks.}
Each panel shows the input image (upper left), and mean attention maps over: (i) all heads (lower left), (ii) the top-20 \Ours{}s (upper right), and (iii) all heads excluding \Ours{}s (lower right).
Green boxes indicate the ground-truth target regions.
  }
  \label{fig:quali_vqa}
\end{figure}

Figure~\ref{fig:quali_vqa} shows analogous visualizations for visual question-answering (VQA) benchmarks: VAW~\citep{pham2021vaw} and Spatial457~\citep{wang2025spatial457}.
As in the grounding case, attention on the visual evidence is preserved when only \Ours{}s are kept but lost when they are masked: the model's answer remains fluent but becomes visually unfaithful, the ``fluent-but-not-factual'' failure mode discussed in Sec.~\ref{sec:functional_specificity}.

\FloatBarrier

\newpage

\section{Application: Direct Preference Optimization Utilizing Visual Retrieval Heads}
\label{app:dpo}
\FloatBarrier

\FloatBarrier

While the main paper focuses on \emph{discovering} VRHs and characterizing their properties, the heads we identify are also actionable: they can be used as targets for downstream optimization. Specifically, we follow the recipe of Ma et al.~\cite{ma2026retmask}, who propose optimizing retrieval heads via DPO~\cite{raf2023dpo} using positive/negative policies obtained from masked versus unmasked models.

\subsection{Method}
We adapt their recipe to the visual setting:
\begin{itemize}[leftmargin=*, itemsep=2pt, topsep=2pt]
\item \textbf{Positive policy} $\pi^+$: the original (unmasked) VLM.
\item \textbf{Negative policy} $\pi^-$: the same VLM with the top-20 VRHs masked.
\item \textbf{DPO data}: for each grounding example $s=(I_s, q_s )\in\mathcal{D}_{\text{train}}$, we sample $y_w\sim\pi^+(\cdot\mid I_s,q_s)$ as the chosen response and $y_l\sim\pi^-(\cdot\mid I_s,q_s)$ as the rejected response.
\item \textbf{Training}: we fine-tune the unmasked VLM on these pairs with the standard DPO loss~\cite{raf2023dpo} with full parameter update,  lr$=2e-7$, for 4 epochs on 6,000 pairs from RefCOCO/+/g train split.
\end{itemize}

As a control, we train a separate model with random-head-masked negatives, sampling 20 random heads (excluding first-layer heads) for each example.

The intuition mirrors the LLM case: the negative policy concentrates exactly the failure mode we want to suppress (visually-unfaithful grounding driven by ablated VRHs), so DPO sharpens the model in the direction of stronger reliance on those same heads.

\subsection{Results}

\begin{figure}[!h]
  \centering
  \includegraphics[width=0.85\linewidth]{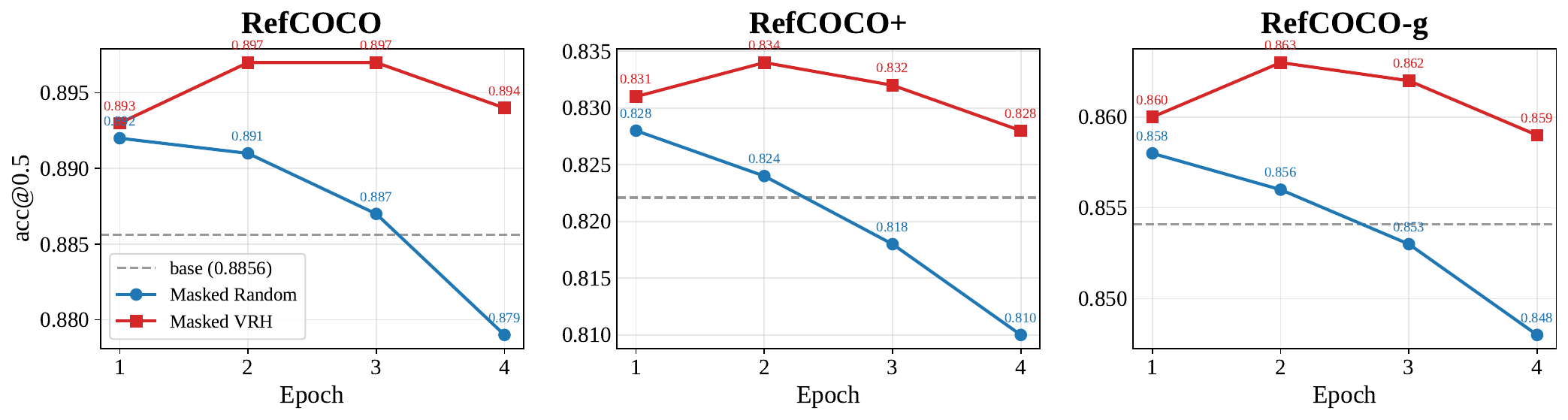}
  \caption{
\textbf{Grounding accuracy comparisons with DPO under varying negative policies.}
We compare two DPO variants that use either random-head dropping or VRH dropping to construct negative responses.
Using VRH dropping as the negative policy yields stronger downstream improvements than random-head dropping, suggesting that perturbing causally relevant heads provides a more informative preference signal for optimization.
  }
  \label{fig:dpo_comparison}
\end{figure}

As shown in Fig.~\ref{fig:dpo_comparison}, we observe modest but consistent gains, with an average improvement of 0.78 percentage-points across three visual grounding benchmarks~\citep{yu2016refcoco,mao2016refcocog}.
This suggests that VRHs are not merely an analytic abstraction, but a mechanistic substrate that can be directly optimized.
We view this as an initial application; using VRHs for routing decisions, KV-cache compression analogous to the LLM setting~\citep{wu2025retrieval}, or test-time intervention are promising directions for future work.

\section{Full Quantitative Results}
\label{app:fullquant}

The main paper reports averaged results over RefCOCO/+/g~\citep{yu2016refcoco,mao2016refcocog}.
Here, we provide the full per-dataset breakdown and per-seed variability.

\subsection{Per-dataset Visual Grounding Results under Head Masking}
\label{app:fullquant:refcoco}

Table~\ref{app:tab:refcoco-perdataset} reports visual grounding accuracy separately on five referring-expression visual grounding benchmarks: RefCOCO, RefCOCO+, RefCOCOg, Toloka, and RefSpatialBench~\citep{yu2016refcoco,mao2016refcocog,ustalov2023toloka,zhou2025roborefer}, for all four VLMs and $k \in \{0, 1, 3, 5, 10, 20\}$ under VRH masking and random-head masking.

\begin{table}[h]
\centering
\caption{\textbf{Per-dataset visual grounding accuracy.}}
\label{app:tab:refcoco-perdataset}
\vspace{\baselineskip}
\small
\begin{tabular}{l l c c c c c c}
\toprule
Model & Method & $k{=}0$ & $k{=}1$ & $k{=}3$ & $k{=}5$ & $k{=}10$ & $k{=}20$ \\
\midrule
\multicolumn{8}{l}{\textbf{RefCOCO~\cite{yu2016refcoco}}} \\
\midrule
Qwen2.5-VL-7B & Random & 86.2  & 86.2 & 86.4 & 86.2 & 85.9 & 86.1
\\
& VRH & 86.2 & 85.1 & 83.5 & 79.3 & 9.7 & 7.5 \\
Qwen3-VL-8B & Random & 91.3 & 91.3 & 91.2 & 91.3 & 91.1 & 91.1 \\ & \Ours{} & 91.3 & 91.0 & 90.4 & 84.1 & 33.4 & 20.4 \\
InternVL3-8B & Random & 93.6 & 93.6 & 93.7 & 93.3 & 92.9 & 92.5 \\
 & \Ours{} & 93.6 & 93.6 & 89.0 & 50.7 & 12.4 & 8.8 \\
InternVL3.5-8B & Random & 90.5 & 90.5 & 90.5 & 90.5 & 90.0 & 88.9 \\
 & \Ours{} & 90.5 & 90.0 & 88.4 & 83.6 & 64.5 & 32.1 \\
\midrule
\midrule
{\textbf{RefCOCOg~\cite{mao2016refcocog}}} \\
\midrule
Qwen2.5-VL-7B & Random & 82.8 & 82.9 & 82.9 & 83.1 & 82.7 & 82.5 \\
 & \Ours{} & 82.8 & 82.5 & 79.9 & 76.1 & 11.8 & 9.2 \\
Qwen3-VL-8B & Random & 87.7 & 87.8 & 87.7 & 87.2 & 87.4 & 87.3 \\
 & \Ours{} & 87.7 & 87.5 & 86.6 & 81.8 & 30.6 & 23.5 \\
InternVL3-8B & Random & 90.0 & 89.9 & 90.0 & 89.8 & 89.4 & 88.9 \\
 & \Ours{} & 90.0 & 90.0 & 62.4 & 56.1 & 17.0 & 12.1 \\
InternVL3.5-8B & Random & 86.8 & 87.0 & 86.9 & 86.9 & 86.8 & 85.7 \\
& \Ours{} & 86.8 & 86.1 & 84.7 & 79.5 & 56.3 & 32.1 \\
\midrule
\midrule
{\textbf{RefCOCO+~\cite{yu2016refcoco}}} \\
\midrule
Qwen2.5-VL-7B & Random & 78.5 & 78.4 & 78.3 & 78.7 & 78.0 & 78.3 \\
 & \Ours{} & 78.5 & 77.9 & 75.3 & 71.5 & 9.4 & 7.5 \\
Qwen3-VL-8B & Random & 84.2 & 84.2 & 84.2 & 84.0 & 83.7 & 83.8 \\
 & \Ours{} & 84.2 & 83.9 & 83.0 & 78.2 & 27.8 & 17.0 \\
InternVL3-8B & Random & 86.8 & 86.8 & 87.0 & 86.8 & 86.2 & 86.0 \\
 & \Ours{} & 86.8 & 84.4 & 81.5 & 45.4 & 11.2 & 8.0 \\
InternVL3.5-8B & Random & 83.4 & 83.3 & 83.4 & 83.4 & 83.4 & 82.2 \\
 & \Ours{} & 83.4 & 82.4 & 81.0 & 75.1 & 52.2 & 25.4 \\
\midrule
\midrule
{\textbf{Toloka~\cite{ustalov2023toloka}}} \\
\midrule
Qwen2.5-VL-7B & Random & 67.4 & 67.2 & 67.4 & 66.9 & 67.6 & 67.2 \\
& \Ours{} & 67.4 & 65.0 & 53.0 & 7.0 & 2.2 & 1.1 \\
Qwen3-VL-8B & Random & 77.7 & 77.3 & 77.7 & 77.6 & 76.8 & 76.5 \\
& \Ours{} & 77.7 & 77.4 & 76.2 & 48.0 & 15.6 & 5.0 \\
InternVL3-8B & Random & 81.2 & 81.2 & 82.0 & 81.8 & 80.7 & 80.2 \\
& \Ours{} & 81.2 & 80.4 & 77.2 & 29.9 & 6.0 & 1.6 \\
InternVL3.5-8B & Random & 67.0 & 66.9 & 66.2 & 66.7 & 67.0 & 61.3 \\
& \Ours{} & 67.0 & 66.3 & 64.0 & 59.9 & 27.0 & 10.3 \\
\midrule
\midrule
{\textbf{RefSpatialBench}~\cite{zhou2025roborefer}} \\
\midrule
Qwen2.5-VL-7B & Random & 43.0 & 43.0 & 37.0 & 42.0 & 40.0 & 36.0 \\
& \Ours{} & 43.0 & 35.0 & 27.0 & 12.0 & 5.0 & 0.0 \\
Qwen3-VL-8B & Random & 47.0 & 46.0 & 49.0 & 45.0 & 46.0 & 46.0 \\
& \Ours{} & 47.0 & 51.0 & 51.0 & 33.0 & 8.0 & 6.0 \\
InternVL3-8B & Random & 39.0 & 37.0 & 39.0 & 40.0 & 42.0 & 40.0 \\
& \Ours{} & 39.0 & 37.0 & 37.0 & 27.0 & 6.0 & 6.0 \\
InternVL3.5-8B & Random & 44.0 & 41.0 & 44.0 & 42.0 & 42.0 & 34.0 \\
& \Ours{} & 44.0 & 40.0 & 24.0 & 20.0 & 11.0 & 5.0 \\

\bottomrule
\end{tabular}
\end{table}
\FloatBarrier

\section{Detected Lists of Visual Retrieval Heads}
\label{app:headlist}

For reproducibility, we list the top-20 VRH (layer, head) indices identified by our scoring method for each (VLM, grounding-benchmark) pair. The lists are ordered by descending head score $\bar\Theta_{l,h}$.

\subsection[Top-20 VRHs On RefCOCO]{Top-20 VRHs On RefCOCO~\cite{yu2016refcoco}}

\vspace{-\baselineskip}

\begin{table}[h]
\centering
\caption{Top-20 VRHs detected on RefCOCO~\cite{yu2016refcoco}. Each entry is $(l,h)$.}
\label{app:tab:headlist-refcoco}
\small
\begin{tabular}{l p{0.8\textwidth}}
\toprule
Model & Top-20 VRH (layer, head) 
\\
\midrule
Qwen2.5-VL-7B   & (19,17), (16,0), (18,16), (16,20), (19,22), (19,20), (16,1), (19,23), (19,15), (17,18), (16,7), (14,11), (14,16), (16,6), (16,17), (19,26), (19,19), (14,23), (17,14), (18,15)
\\
Qwen3-VL-8B     & (20,15), (21,10), (18,19), (18,15), (21,11), (21,18), (17,27), (17,4), (19,3), (22,4), (21,8), (20,8), (18,5), (23,13), (20,17), (20,18), (23,30), (21,25), (20,16), (23,10)
\\
InternVL3-8B    & (16,20), (16,0), (19,22), (16,1), (19,23), (19,17), (19,20), (16,6), (14,11), (19,15), (16,4), (17,24), (18,16), (16,16), (16,17), (20,21), (16,7), (19,19), (14,16), (14,8) 
\\
InternVL3.5-8B  & (21,11), (20,15), (21,10), (19,3), (18,15), (17,4), (21,18), (18,19), (17,27), (20,17), (24,13), (21,8), (20,18), (20,16), (21,25), (20,8), (18,5), (22,4), (18,21), (23,30) \\
\bottomrule
\end{tabular}
\end{table}
\FloatBarrier

\subsection[Top-20 VRHs On RefCOCO+]{Top-20 VRHs On RefCOCO+~\cite{yu2016refcoco}}

\vspace{-\baselineskip}

\begin{table}[h]
\centering
\caption{Top-20 VRHs detected on RefCOCO+~\cite{yu2016refcoco}. Each entry is $(l,h)$.}
\label{app:tab:headlist-refcocoplus}
\small
\begin{tabular}{l p{0.8\textwidth}}
\toprule
Model & Top-20 VRH (layer, head) \\
\midrule
Qwen2.5-VL-7B   & (19,17), (16,0), (16,20), (18,16), (19,22), (19,20), (16,1), (19,15), (19,23), (17,18), (14,11), (16,7), (14,16), (16,6), (16,17), (19,19), (19,26), (17,24), (14,23), (17,14) \\
Qwen3-VL-8B     & (20,15), (21,10), (18,19), (18,15), (17,27), (21,11), (21,18), (17,4), (19,3), (20,8), (22,4), (21,8), (20,17), (20,18), (23,13), (21,25), (18,5), (23,30), (20,16), (23,10) \\
InternVL3-8B    & (16,0), (16,20), (19,22), (19,23), (16,1), (19,17), (19,20), (16,6), (19,15), (14,11), (16,4), (17,24), (18,16), (20,21), (16,7), (16,16), (16,17), (14,8), (14,16), (19,19) \\
InternVL3.5-8B  & (21,11), (21,10), (20,15), (19,3), (18,15), (21,18), (17,4), (18,19), (17,27), (20,17), (21,8), (20,18), (20,16), (24,13), (20,8), (22,4), (21,25), (17,26), (23,30), (23,10) \\
\bottomrule
\end{tabular}
\end{table}
\FloatBarrier

\subsection[Top-20 VRHs On RefCOCOg]{Top-20 VRHs On RefCOCOg~\cite{mao2016refcocog}}

\vspace{-\baselineskip}

\begin{table}[h]
\centering
\caption{Top-20 VRHs detected on RefCOCOg~\cite{mao2016refcocog}. Each entry is $(l,h)$.}
\label{app:tab:headlist-refcocog}
\small
\begin{tabular}{l p{0.8\textwidth}}
\toprule
Model & Top-20 VRH (layer, head) \\
\midrule
Qwen2.5-VL-7B   & (19,17), (16,0), (16,20), (19,22), (18,16), (16,1), (19,20), (19,23), (19,15), (14,11), (17,18), (16,7), (14,16), (16,6), (16,17), (19,26), (14,23), (19,19), (17,14), (18,15) \\
Qwen3-VL-8B     & (20,15), (18,19), (21,10), (17,27), (18,15), (21,18), (21,11), (17,4), (19,3), (22,4), (21,8), (20,8), (18,5), (20,17), (20,18), (21,25), (23,30), (23,13), (20,16), (17,24) \\
InternVL3-8B    & (16,20), (16,0), (16,1), (19,22), (19,17), (19,23), (14,11), (16,6), (19,20), (16,4), (19,15), (18,16), (16,17), (16,16), (17,24), (14,8), (14,16), (16,7), (20,21), (14,14) \\
InternVL3.5-8B  & (21,11), (20,15), (21,10), (18,15), (19,3), (17,4), (21,18), (18,19), (17,27), (20,17), (20,18), (20,16), (21,8), (24,13), (19,13), (20,8), (18,21), (17,26), (22,4), (18,5) \\
\bottomrule
\end{tabular}
\end{table}
\FloatBarrier

\newpage

\subsection[Top-20 VRHs On Toloka]{Top-20 VRHs On Toloka~\cite{ustalov2023toloka}}

\vspace{-\baselineskip}

\begin{table}[h]
\centering
\caption{Top-20 VRHs detected on Toloka~\cite{ustalov2023toloka}. Each entry is $(l,h)$.}
\label{app:tab:headlist-toloka}
\small
\begin{tabular}{l p{0.8\textwidth}}
\toprule
Model & Top-20 VRH (layer, head) \\
\midrule
Qwen2.5-VL-7B   & (19,17), (16,0), (19,22), (16,20), (19,15), (19,20), (19,23), (16,1), (18,16), (17,18), (19,19), (16,6), (19,26), (14,11), (19,25), (16,17), (16,16), (16,18), (16,4), (20,21) \\
Qwen3-VL-8B     & (21,10), (20,15), (18,19), (21,11), (17,4), (18,15), (23,30), (21,8), (23,13), (22,4), (21,18), (20,17), (17,27), (23,10), (20,16), (20,8), (20,18), (19,3), (24,16), (21,31) \\
InternVL3-8B    & (16,20), (19,17), (16,1), (19,20), (16,0), (19,15), (19,22), (16,17), (14,11), (19,23), (18,16), (16,16), (16,6), (20,21), (14,8), (19,19), (16,7), (17,24), (16,18), (21,25) \\
InternVL3.5-8B  & (18,19), (21,11), (17,27), (21,10), (20,15), (18,15), (20,17), (17,4), (24,13), (23,30), (21,8), (19,3), (20,16), (21,18), (20,8), (23,10), (20,23), (18,21), (19,13), (21,9) \\
\bottomrule
\end{tabular}
\end{table}
\FloatBarrier

\subsection[Top-20 VRHs On RefSpatialBench]{Top-20 VRHs On RefSpatialBench~\cite{zhou2025roborefer}}

\vspace{-\baselineskip}

\begin{table}[h]
\centering
\caption{Top-20 VRHs detected on RefSpatialBench~\cite{zhou2025roborefer}. Each entry is $(l,h)$.}
\label{app:tab:headlist-refspatial}
\small
\begin{tabular}{l p{0.8\textwidth}}
\toprule
Model & Top-20 VRH (layer, head) \\
\midrule
Qwen2.5-VL-7B   & (19,17), (19,22), (19,20), (19,15), (16,0), (16,20), (19,23), (16,1), (19,19), (18,16), (17,18), (19,26), (20,21), (16,16), (16,6), (16,4), (14,11), (19,25), (16,7), (20,23) \\
Qwen3-VL-8B     & (21,10), (21,11), (20,15), (21,8), (23,30), (17,4), (18,19), (18,15), (21,18), (23,13), (23,10), (22,4), (20,17), (19,3), (20,16), (17,27), (20,8), (23,14), (25,10), (21,31) \\
InternVL3-8B    & (19,17), (16,1), (16,20), (19,22), (19,15), (19,20), (19,23), (16,0), (14,11), (20,21), (17,24), (19,19), (16,6), (18,16), (16,17), (16,7), (16,16), (21,25), (21,5), (16,4) \\
InternVL3.5-8B  & (21,11), (21,10), (21,8), (18,15), (19,3), (18,19), (21,18), (20,17), (17,4), (20,15), (17,27), (20,8), (24,13), (23,30), (20,23), (20,16), (21,25), (23,10), (20,21), (21,9) \\
\bottomrule
\end{tabular}
\end{table}
\FloatBarrier

\section{Experimental Setup}
\label{app:setup}

We provide additional details on the VLMs, benchmarks, and head-masking
procedure used throughout the paper.

\subsection{Vision-Language Models}
\label{app:setup:vlms}

We evaluate our method on four open-source VLMs spanning two model families:
Qwen2.5-VL-7B~\cite{bai2025qwen25vl}, Qwen3-VL-8B~\cite{bai2025qwen3vl}, InternVL3-8B~\cite{zhu2025internvl3},
and InternVL3.5-8B~\cite{wang2025internvl35}. These four models form two same-backbone pairs that differ in vision encoder, projector, and instruction-tuning recipe. This
controlled pairing is what enables the cross-model VRH transfer analysis in
Sec.~\ref{sec:architecturally_shared} of the main paper.

Table~\ref{app:tab:vlm-arch} summarizes the architectural specifications relevant
to our analysis: the number of layers $L$, attention heads per layer $H$, and
total head count $L\!\cdot\!H$ that appears in the denominator of our sparsity
ratios (e.g., ``20 / 784 = 2.6\%'' for Qwen2.5-VL-7B).

\begin{table}[h]
\centering
\caption{Architectural details of the four evaluated VLMs.}
\label{app:tab:vlm-arch}
\small
\setlength{\tabcolsep}{4pt}
\resizebox{\linewidth}{!}{%
\begin{tabular}{lcccccc}
\toprule
Model & \# Layers $L$ & \# Heads/layer $H$ & Total heads & Language backbone & Vision encoder & Projector \\
\midrule
Qwen2.5-VL-7B   & 28 & 28 & 784  & Qwen2.5-7B & Qwen ViT       & MLP merger \\
Qwen3-VL-8B     & 36 & 32 & 1152 & Qwen3-8B   & Qwen ViT       & MLP merger + DeepStack \\
InternVL3-8B    & 28 & 28 & 784  & Qwen2.5-7B & InternViT-300M & 2-layer MLP \\
InternVL3.5-8B  & 36 & 32 & 1152 & Qwen3-8B   & InternViT-300M & 2-layer MLP \\
\bottomrule
\end{tabular}}
\end{table}

\subsection{Benchmarks}
\label{app:setup:benchmarks}

\paragraph{Visual grounding benchmarks (head detection \& main causality evaluation).}
We use five referring-expression visual grounding datasets, grouped into three families:
\begin{itemize}[leftmargin=*, itemsep=2pt, topsep=2pt]
\item \textbf{RefCOCO/+/g}~\cite{yu2016refcoco, mao2016refcocog}: standard COCO-derived referring-expression benchmarks. We evaluate on the validation splits and use  8,811, 3,804, 7,573 examples per split.
\item \textbf{RefSpatialBench}~\cite{zhou2025roborefer}: a referring-expression benchmark emphasizing spatial-relation expressions. We use 100 examples (full set).
\item \textbf{Toloka VQA}~\cite{ustalov2023toloka}: a crowd-sourced visual referring benchmark with diverse expression styles. We use 1705 examples (full set).
\end{itemize}
For all grounding datasets, a prediction is judged correct if $\mathrm{IoU}(\hat B_s, B_s)\ge \tau=0.5$ (consistent with the main paper).

\paragraph{VQA benchmarks (cross-task generalization in Sec.~5.1).}
We further evaluate masked VLMs on seven VQA benchmarks spanning attribute prediction (VAW~\cite{pham2021vaw}, Spatial457~\cite{wang2025spatial457}), spatial reasoning (SpatialRGPT~\cite{cheng2024spatialrgpt},
CV-Bench~\cite{tong2024cvbench}), vision-centric understanding (MMStar~\cite{chen2024mmstar}),
counting (CountBenchQA~\cite{paiss2023countbench}), and visual math (MathVista~\cite{lu2023mathvista}).

\begin{itemize}[leftmargin=*, itemsep=2pt, topsep=2pt]
    \item \textbf{VAW}~\cite{pham2021vaw}: We manually convert VAW into a multiple-choice question (MCQ) format and evaluate on 6,309 samples.
    \item \textbf{Spatial457}~\cite{wang2025spatial457}: We use the L1 (Single Object) portion and evaluate on 670 samples.
    \item \textbf{SpatialRGPT}~\cite{cheng2024spatialrgpt}: We use the quantitative object-size estimation subset and evaluate on 266 samples.
    \item \textbf{CV-Bench}~\cite{tong2024cvbench}: We use the full benchmark and evaluate on 2,638 samples.
    \item \textbf{MMStar}~\cite{chen2024mmstar}: We use the minitest split and evaluate on 1,500 samples.
    \item \textbf{CountBenchQA}~\cite{paiss2023countbench}: We use the full set of converted QA and evaluate on 491 samples.
    \item \textbf{MathVista}~\cite{lu2023mathvista}: We use the MCQ split and evaluate on 1,000 samples.
\end{itemize}

\subsection{Head Masking Procedure}
\label{app:setup:masking}

To evaluate the causal contribution of a head $(l, h)$, we ablate it during the forward pass by  zeroing the attention output of head $(l, h)$ before it is added to the residual stream, leaving all other heads, MLPs, and layer norms untouched.
This is the same masking convention used by Wu et al.~\cite{wu2025retrieval} for text retrieval heads, ensuring our results are directly comparable.

\paragraph{Generation settings.}
All evaluations use \textit{greedy decoding} and the model's default chat template. %
For grounding, the bounding box is parsed from the generated text using the model-specific output format (e.g. \texttt{"bbox\_2d": [x1, y1, x2, y2]} for Qwen-VL families).

\subsection{Compute}
All experiments were conducted on a single NVIDIA RTX A6000 (48~GB VRAM).

\newpage

\section{Ablation on the Combined Query Token Set}
\label{app:union}

Sec.~\ref{subsec:causal_validation} of the main paper ablates $\mathcal{Q}^{\mathrm{in}}_{s}$ and $\mathcal{Q}^{\mathrm{out}}_{s}$ separately.
Here we additionally experiment with a \Ours{} detection setting that uses their union $\mathcal{Q}^{\mathrm{in}}_{s} \cup \mathcal{Q}^{\mathrm{out}}_{s}$, and compare the causal effectiveness of the resulting heads with that of our original \Ours{}s in Tab.~\ref{app:tab:union}.
Because this setting uses a superset of the query tokens compared to our original $\mathcal{Q}^{\mathrm{out}}_{s}$ setting, it achieves comparable causal effectiveness for some models.
However, our original setting still performs better on three models, Qwen3-VL, InternVL3 and InternVL3.5, with a particularly noticeable gap on Qwen3-VL.
These results suggest that our original setting is more robust across models while also being more efficient, as it uses substantially fewer query tokens for head detection.

\begin{table}[h]
\centering
\caption{\textbf{Visual grounding accuracy on RefCOCO~\citep{yu2016refcoco} after masking the top-20 \Ours{}s detected using $\mathcal{Q}^{\mathrm{in}}_{s} \cup \mathcal{Q}^{\mathrm{out}}_{s}$ versus $\mathcal{Q}^{\mathrm{out}}_{s}$ (ours) across four VLMs.}}
\label{app:tab:union}
\vspace{0.5\baselineskip}
\small
\setlength{\tabcolsep}{4pt}
\begin{tabular}{lcccc}
\toprule
& Qwen2.5-VL-7B & Qwen3-VL-8B & InternVL3-8B & InternVL3.5-8B \\
\midrule
Original model & 86.11 & 91.27 & 93.59 & 90.48 \\
w/ Masking \Ours{}s ($\mathcal{Q}^{\mathrm{in}}_{s} \cup \mathcal{Q}^{\mathrm{out}}_{s}$) & 7.14 & 29.85 & 8.98 & 32.94 \\
w/ Masking \Ours{}s ($\mathcal{Q}^{\mathrm{out}}_{s}$; ours) & 7.48 & 20.39 & 8.76 & 32.03 \\
\bottomrule
\end{tabular}
\end{table}
\FloatBarrier

\newpage

\section{Sensitivity to the Number of Masked Heads}
\label{app:ksweep}

Fig.~\ref{fig:score_ablation} of the main paper reports the scoring-variant ablation at $k{=}20$.
Tab.~\ref{app:tab:ksweep} reports the same ablation for $k \in \{5, 10, 20, 50\}$ across the four VLMs, comparing our design choice $(\mathcal{Q}^{\mathrm{out}}_s, \Phi^{\mathrm{sum}}, \Omega^{\mathrm{mean}})$ against variants obtained by flipping one design component at a time and against random-head masking.
At $k{=}5$ the masking effect is relatively weak or inconsistent across models, which makes the relative ranking among scoring variants less conclusive.
Once $k$ increases beyond $5$, our design choice consistently ranks either first or second in terms of causal effect across $k \in \{10, 20, 50\}$.
Notably, masking only the top-10 \Ours{}s already causes substantial degradation across all scoring variants, showing that a small subset of \Ours{}s is sufficient to strongly impair grounding.
We use $k{=}20$ in the main experiments, following the convention established in the retrieval-head literature~\citep{wu2025retrieval,baek2025ocrhead}; Tab.~\ref{app:tab:ksweep} shows that our main conclusion remains largely robust for $k \ge 10$.

\begin{table}[h]
\centering
\caption{\textbf{Scoring ablation for varying numbers of masked heads $k \in \{5, 10, 20, 50\}$ across four VLMs.} We report RefCOCO~\citep{yu2016refcoco} visual grounding accuracy after masking random heads, our detected \Ours{}s, or \Ours{}s obtained by flipping each scoring design choice. \textbf{Bold} and \underline{underline} mark the largest and the second-largest causal effect among the four scoring variants in each row.}
\label{app:tab:ksweep}
\vspace{0.5\baselineskip}
\small
\begin{tabular}{lccccc}
\toprule
& Random heads & \Ours{}s (Ours) & Flipped $\mathcal{Q}_s$ & Flipped $\Phi$ & Flipped $\Omega$ \\
\midrule
\multicolumn{6}{l}{\textbf{(a) Qwen2.5-VL-7B}} \\
\midrule
$k{=}0$ (original model) & 86.11 & 86.11 & 86.11 & 86.11 & 86.11 \\
$k{=}5$ & 86.16 & 79.29 & \textbf{56.08} & 63.28 & \underline{63.14} \\
$k{=}10$ & 85.79 & \underline{9.63} & 10.52 & \textbf{8.87} & 9.63 \\
$k{=}20$ & 85.97 & \textbf{7.48} & \underline{7.68} & 8.13 & 8.01 \\
$k{=}50$ & 84.67 & \textbf{6.63} & 9.85 & \underline{7.21} & 8.70 \\
\midrule
\multicolumn{6}{l}{\textbf{(b) Qwen3-VL-8B}} \\
\midrule
$k{=}0$ & 91.27 & 91.27 & 91.27 & 91.27 & 91.27 \\
$k{=}5$ & 91.26 & 84.14 & \underline{79.90} & \textbf{57.03} & 84.24 \\
$k{=}10$ & 91.12 & \underline{33.42} & 80.47 & 33.60 & \textbf{32.63} \\
$k{=}20$ & 91.09 & \textbf{20.39} & 51.10 & 33.61 & \underline{21.96} \\
$k{=}50$ & 89.97 & \underline{9.29} & \textbf{7.20} & 14.17 & 13.11 \\
\midrule
\multicolumn{6}{l}{\textbf{(c) InternVL3-8B}} \\
\midrule
$k{=}0$ & 93.59 & 93.59 & 93.59 & 93.59 & 93.59 \\
$k{=}5$ & 93.28 & \textbf{50.62} & 72.06 & \underline{50.62} & 56.16 \\
$k{=}10$ & 92.90 & \textbf{12.39} & 39.05 & \underline{14.20} & 14.20 \\
$k{=}20$ & 92.46 & \underline{8.76} & 15.83 & \textbf{7.89} & 9.93 \\
$k{=}50$ & 90.98 & \underline{8.31} & 13.90 & \textbf{7.35} & 8.87 \\
\midrule
\multicolumn{6}{l}{\textbf{(d) InternVL3.5-8B}} \\
\midrule
$k{=}0$ & 90.48 & 90.48 & 90.48 & 90.48 & 90.48 \\
$k{=}5$ & 90.45 & \underline{83.52} & 86.94 & \textbf{82.22} & 83.52 \\
$k{=}10$ & 89.95 & \textbf{64.48} & 71.50 & 69.83 & \underline{64.48} \\
$k{=}20$ & 88.86 & \textbf{32.03} & 45.66 & 63.96 & \underline{37.56} \\
$k{=}50$ & 88.09 & \underline{16.98} & \textbf{10.42} & 23.49 & 19.73 \\
\bottomrule
\end{tabular}
\end{table}
\FloatBarrier

\newpage

\section{Head Detection and Evaluation Splits}
\label{app:split}

\paragraph{Split protocol.}
For each grounding benchmark we use a fixed detection set for head scoring and evaluate the masked models on the remaining samples.
On RefCOCO~\citep{yu2016refcoco} the detection set consists of $200$ samples drawn from the validation split and the evaluation set is the remaining $8{,}611$ samples, so the two sets are disjoint by construction.
In our original RefCOCO setup we used $200$ samples for head detection and the full set of $8{,}811$ samples for evaluation.
Because the detection set constitutes only a small portion of the full dataset, we did not exclude it from evaluation, as we anticipated that it would have a minimal effect on the results.
Tab.~\ref{app:tab:split} reports causal validation results both with and without these $200$ head-scoring samples; the results do not meaningfully differ.
We nevertheless use disjoint head detection and evaluation sets as the default protocol.

We also note that Tab.~\ref{tab:downstream} of the main paper includes cross-task causal evaluations in which heads detected using $200$ RefCOCO samples are evaluated on separate, non-visual-grounding benchmarks, effectively alleviating concerns about similarity between the head scoring and evaluation sets.

\begin{table}[h]
\centering
\caption{\textbf{Visual grounding accuracy on RefCOCO~\citep{yu2016refcoco} after masking the top-20 \Ours{}s, evaluated with and without the $200$ head-detection samples.}}
\label{app:tab:split}
\vspace{0.5\baselineskip}
\small
\setlength{\tabcolsep}{3pt}
\begin{tabular}{lcccc}
\toprule
Evaluation set & Qwen2.5-VL-7B & Qwen3-VL-8B & InternVL3-8B & InternVL3.5-8B \\
\midrule
w/ head detection samples ($n{=}8{,}811$)  & 7.46 & 20.44 & 8.76 & 32.14 \\
w/o head detection samples ($n{=}8{,}611$) & 7.48 & 20.39 & 8.76 & 32.03 \\
\bottomrule
\end{tabular}
\end{table}
\FloatBarrier

\section{Detection Cost and Fused Attention}
\label{app:cost}

\Ours{} detection requires explicitly extracting attention weights over visual tokens, so its cost grows with the number of visual tokens.
Tab.~\ref{app:tab:cost} reports the peak memory usage and computation time of \Ours{} detection while varying the number of visual tokens by changing the input resolution or the number of input images.
To examine how memory and computational costs interact with the use of fused-attention kernels, we report results using both eager and fused attention.
As the number of visual tokens increases from $256$ to $512$ and $3{,}328$, both memory usage and computation time increase, with the growth being substantially more pronounced under eager attention.
In particular, at $3{,}328$ visual tokens, eager attention requires $34.59$~GiB and $4.86$ s/sample, compared with $16.01$~GiB and $2.31$ s/sample using fused attention.
These results show that fused-attention kernels substantially mitigate the memory and computational overhead associated with scaling to larger numbers of visual tokens.
This is because eager attention materializes dense attention maps whose memory grows approximately quadratically with sequence length, whereas fused attention avoids retaining the full maps by recomputing only the query rows needed for \Ours{} scoring.

\begin{table}[h]
\centering
\caption{\textbf{Peak memory usage and computation time of \Ours{} detection with varying numbers of visual tokens.} We use InternVL3-8B~\citep{zhu2025internvl3} with dynamic tiling as the base model and conduct all measurements on a single 48~GB A6000 GPU.}
\label{app:tab:cost}
\vspace{0.5\baselineskip}
\small
\setlength{\tabcolsep}{5pt}
\begin{tabular}{lccc}
\toprule
Setting & \# Visual tokens & \begin{tabular}[c]{@{}c@{}}Peak memory (GiB)\\ (Eager / Fused Att.)\end{tabular} & \begin{tabular}[c]{@{}c@{}}Time (s/sample)\\ (Eager / Fused Att.)\end{tabular} \\
\midrule
Low resolution ($448{\times}448$)              & 256   & 15.09 / 14.92 & $1.77 \pm 0.47$ / $1.41 \pm 0.21$ \\
Low resolution ($448{\times}448$), 2 images    & 512   & 15.55 / 15.01 & $2.33 \pm 0.99$ / $1.56 \pm 0.34$ \\
High resolution ($1792{\times}1344$)           & 3,328 & 34.59 / 16.01 & $4.86 \pm 0.77$ / $2.31 \pm 0.29$ \\
\bottomrule
\end{tabular}
\end{table}
\FloatBarrier

\newpage

\section{Evaluated VLMs: Protocols and Per-Model Results}
\label{app:models}

This section provides the protocol and the exact values behind Fig.~\ref{fig:allmodels} and Fig.~\ref{fig:same_backbone_consistency} of the main paper.

\paragraph{Protocol.}
The four VLMs studied throughout the paper follow the split of Sec.~\ref{app:split}.
For the seven further VLMs, \Ours{}s are detected on a $200$-sample RefCOCO~\citep{yu2016refcoco} detection set using the same scoring function, and the masked models are evaluated on a disjoint $500$-sample evaluation set of the same benchmark.
The random-head control masks the same number of heads, $k{=}20$, drawn outside the detected set.
Tab.~\ref{tab:allmodels} lists their exact accuracies as plotted in Fig.~\ref{fig:allmodels}.

\begin{table}[!h]
\centering
\caption{
\textbf{Exact values plotted in Fig.~\ref{fig:allmodels}} for the seven VLMs introduced in addition to the four studied throughout the paper.
Visual grounding accuracy on RefCOCO~\cite{yu2016refcoco} after masking the top-20 \Ours{}s versus the same number of random heads.
GLM-4.6V-Flash and Phi-4-Reasoning-Vision were released after the main experiments of this work were completed.
}
\label{tab:allmodels}
\vspace{0.5\baselineskip}
\resizebox{\linewidth}{!}{\begin{tabular}{lccccccc}
\toprule
Condition
& Magma-8B~\cite{yang2025magma}
& \begin{tabular}[c]{@{}c@{}}LLaVA-NeXT-\\ Llama3-8B~\cite{li2024llavanext}\end{tabular}
& \begin{tabular}[c]{@{}c@{}}PaliGemma2-\\ 10B~\cite{steiner2024paligemma2}\end{tabular}
& \begin{tabular}[c]{@{}c@{}}DeepSeek-VL2-\\ Small~\cite{wu2024deepseekvl2}\end{tabular}
& \begin{tabular}[c]{@{}c@{}}InternVL2.5-\\ 8B~\cite{chen2024internvl25}\end{tabular}
& \begin{tabular}[c]{@{}c@{}}GLM-4.6V-\\ Flash~\cite{glm2025glm4v}\end{tabular}
& \begin{tabular}[c]{@{}c@{}}Phi-4-Reasoning-\\ Vision~\cite{abouelenin2025phi4rv}\end{tabular} \\
Language backbone & Llama-3 & Llama-3 & Gemma 2 & DeepSeekMoE & InternLM2.5 & GLM & Phi \\
\midrule
Baseline ($k{=}0$) & 68.8 & 85.8 & 73.6 & 93.0 & 90.0 & 86.4 & 74.2 \\
Random ($k{=}20$) & 64.0 & 77.8 & 73.2 & 91.6 & 86.8 & 85.8 & 73.8 \\
\Ours{} ($k{=}20$) & \textbf{7.8} & \textbf{5.0} & \textbf{11.6} & \textbf{21.4} & \textbf{11.4} & \textbf{10.0} & \textbf{18.0} \\
\bottomrule
\end{tabular}}
\vspace{-0.5\baselineskip}
\end{table}

\paragraph{Cross-model transfer for the Llama-3 pair.}
Magma-8B~\citep{yang2025magma} and LLaVA-NeXT-Llama3-8B~\citep{li2024llavanext} share Llama-3-based language backbones, making them a third same-backbone pair for the analysis of Sec.~\ref{sec:architecturally_shared}; their score maps and the full $k$-sweep are plotted in Fig.~\ref{fig:same_backbone_consistency} of the main paper.
As shown in Tab.~\ref{app:tab:pair-cross}, cross-masking \Ours{}s causes a significant drop in visual grounding accuracy in both directions, supporting the claim that \Ours{}s are transferable across models with the same LLM backbone in non-Qwen families as well.

\begin{table}[h]
\centering
\caption{\textbf{Cross-model transferability of \Ours{}s between Magma-8B~\citep{yang2025magma} and LLaVA-NeXT-Llama3-8B~\citep{li2024llavanext}.} We report RefCOCO~\citep{yu2016refcoco} visual grounding accuracy before and after masking the top-20 \Ours{}s identified in the other model.}
\label{app:tab:pair-cross}
\vspace{0.5\baselineskip}
\small
\begin{tabular}{lcc}
\toprule
& Magma-8B & LLaVA-NeXT-Llama3-8B \\
\midrule
Original model & 68.8 & 85.8 \\
w/ Masking own top-20 \Ours{}s & 7.8 & 5.0 \\
w/ Cross-masking \Ours{}s & 0.0 & 5.8 \\
w/ Masking random heads & 64.0 & 77.8 \\
\bottomrule
\end{tabular}
\end{table}
\FloatBarrier

\FloatBarrier


{
\small
\begin{thebibliography}{10}

\bibitem{wu2025retrieval}
Wenhao Wu, Yizhong Wang, Guangxuan Xiao, Hao Peng, and Yao Fu.
\newblock Retrieval head mechanistically explains long-context factuality.
\newblock In {\em ICLR}, 2025.

\bibitem{baek2025ocrhead}
Ingeol Baek, Hwan Chang, Sunghyun Ryu, and Hwanhee Lee.
\newblock How do large vision-language models see text in image? unveiling the
  distinctive role of ocr heads.
\newblock In {\em EMNLP}, 2025.

\bibitem{pei2026vera}
Rongcan Pei, Huan Li, Fang Guo, and Qi~Zhu.
\newblock Vera: Identifying and leveraging visual evidence retrieval heads in
  long-context understanding.
\newblock {\em arXiv preprint arXiv:2602.10146}, 2026.

\bibitem{kang2024locheads}
Seil Kang, Jinyeong Kim, Junhyeok Kim, and Seong~Jae Hwang.
\newblock Your large vision-language model only needs a few attention heads for
  visual grounding.
\newblock In {\em CVPR}, 2025.

\bibitem{yu2016refcoco}
Licheng Yu, Patrick Poirson, Shan Yang, Alexander~C Berg, and Tamara~L Berg.
\newblock Modeling context in referring expressions.
\newblock In {\em ECCV}, 2016.

\bibitem{bai2025qwen25vl}
Shuai Bai, Keqin Chen, Xuejing Liu, Jialin Wang, Wenbin Ge, Sibo Song, Kai
  Dang, Peng Wang, Shijie Wang, Jun Tang, et~al.
\newblock Qwen2.5-{VL} technical report.
\newblock {\em arXiv preprint arXiv:2502.13923}, 2025.

\bibitem{bai2025qwen3vl}
Shuai Bai, Yuxuan Cai, Ruizhe Chen, et~al.
\newblock Qwen3-vl technical report.
\newblock {\em arXiv preprint arXiv:2511.21631}, 2025.

\bibitem{zhu2025internvl3}
Jinguo Zhu, Weiyun Wang, Zhe Chen, Zhaoyang Liu, Shenglong Ye, Lixin Gu, Yuchen
  Duan, Hao Tian, Weijie Su, Jie Shao, et~al.
\newblock Internvl3: Exploring advanced training and test-time recipes for
  open-source multimodal models.
\newblock {\em arXiv preprint arXiv:2504.10479}, 2025.

\bibitem{wang2025internvl35}
Weiyun Wang, Zhangwei Gao, Lixin Gu, Hengjun Pu, Long Cui, Xingguang Wei,
  Zhaoyang Liu, Linglin Jing, Shenglong Ye, Jie Shao, et~al.
\newblock Internvl3.5: Advancing open-source multimodal models in versatility,
  reasoning, and efficiency.
\newblock {\em arXiv preprint arXiv:2508.18265}, 2025.

\bibitem{mao2016refcocog}
Junhua Mao, Jonathan Huang, Alexander Toshev, Oana Camburu, Alan Yuille, and
  Kevin Murphy.
\newblock Generation and comprehension of unambiguous object descriptions.
\newblock In {\em CVPR}, 2016.

\bibitem{zhou2025roborefer}
Enshen Zhou, Jingkun An, Cheng Chi, Yi~Han, Shanyu Rong, Chi Zhang, Pengwei
  Wang, Zhongyuan Wang, Tiejun Huang, Lu~Sheng, et~al.
\newblock Roborefer: Towards spatial referring with reasoning in
  vision-language models for robotics.
\newblock In {\em NeurIPS}, 2025.

\bibitem{ustalov2023toloka}
Dmitry Ustalov, Nikita Pavlichenko, Sergey Koshelev, Daniil Likhobaba, and
  Alisa Smirnova.
\newblock Toloka visual question answering benchmark.
\newblock {\em arXiv preprint arXiv:2309.16511}, 2023.

\bibitem{pham2021vaw}
Khoi Pham, Kushal Kafle, Zhe Lin, Zhihong Ding, Scott Cohen, Quan Tran, and
  Abhinav Shrivastava.
\newblock Learning to predict visual attributes in the wild.
\newblock In {\em CVPR}, 2021.

\bibitem{wang2025spatial457}
Xingrui Wang, Wufei Ma, Tiezheng Zhang, Celso~M de~Melo, Jieneng Chen, and Alan
  Yuille.
\newblock Spatial457: A diagnostic benchmark for 6d spatial reasoning of large
  mutimodal models.
\newblock In {\em CVPR}, 2025.

\bibitem{cheng2024spatialrgpt}
An-Chieh Cheng, Hongxu Yin, Yang Fu, Qiushan Guo, Ruihan Yang, Jan Kautz,
  Xiaolong Wang, and Sifei Liu.
\newblock Spatialrgpt: Grounded spatial reasoning in vision-language models.
\newblock In {\em NeurIPS}, 2024.

\bibitem{tong2024cvbench}
Shengbang Tong, Ellis Brown, Penghao Wu, Sanghyun Woo, Manoj Middepogu, Sai~C
  Akula, Jihan Yang, Shusheng Yang, Adithya Iyer, Xichen Pan, et~al.
\newblock Cambrian-1: A fully open, vision-centric exploration of multimodal
  llms.
\newblock In {\em NeurIPS}, 2024.

\bibitem{chen2024mmstar}
Lin Chen, Jinsong Li, Xiaoyi Dong, Pan Zhang, Yuhang Zang, Zehui Chen, Haodong
  Duan, Jiaqi Wang, Yu~Qiao, Dahua Lin, et~al.
\newblock Are we on the right way for evaluating large vision-language models?
\newblock In {\em NeurIPS}, 2024.

\bibitem{paiss2023countbench}
Roni Paiss, Ariel Ephrat, Omer Tov, Shiran Zada, Inbar Mosseri, Michal Irani,
  and Tali Dekel.
\newblock Teaching clip to count to ten.
\newblock In {\em ICCV}, 2023.

\bibitem{lu2023mathvista}
Pan Lu, Hritik Bansal, Tony Xia, Jiacheng Liu, Chunyuan Li, Hannaneh
  Hajishirzi, Hao Cheng, Kai-Wei Chang, Michel Galley, and Jianfeng Gao.
\newblock Mathvista: Evaluating mathematical reasoning of foundation models in
  visual contexts.
\newblock In {\em ICLR}, 2024.

\bibitem{zhang2025qrhead}
Wuwei Zhang, Fangcong Yin, Howard Yen, Danqi Chen, and Xi~Ye.
\newblock Query-focused retrieval heads improve long-context reasoning and
  re-ranking.
\newblock In {\em EMNLP}, 2025.

\bibitem{goldowskydill2023pathpatching}
Nicholas Goldowsky-Dill, Chris MacLeod, Lucas Sato, and Aryaman Arora.
\newblock Localizing model behavior with path patching.
\newblock {\em arXiv preprint arXiv:2304.05969}, 2023.

\bibitem{li2026causaltracing}
Qiming Li, Zekai Ye, Xiaocheng Feng, Weihong Zhong, Weitao Ma, and Xiachong
  Feng.
\newblock Causal tracing of object representations in large vision language
  models: Mechanistic interpretability and hallucination mitigation.
\newblock In {\em AAAI}, 2026.

\bibitem{wang2025cdviews}
Fengyun Wang, Sicheng Yu, Jiawei Wu, Jinhui Tang, Hanwang Zhang, and Qianru
  Sun.
\newblock 3d question answering via only 2d vision-language models.
\newblock In {\em ICML}, 2025.

\bibitem{ge2024model}
Suyu Ge, Yunan Zhang, Liyuan Liu, Minjia Zhang, Jiawei Han, and Jianfeng Gao.
\newblock Model tells you what to discard: Adaptive {KV} cache compression for
  {LLM}s.
\newblock In {\em ICLR}, 2024.

\bibitem{ma2026retmask}
Youmi Ma and Naoaki Okazaki.
\newblock From interpretability to performance: Optimizing retrieval heads for
  long-context language models.
\newblock In {\em Findings of ACL}, 2026.

\bibitem{raf2023dpo}
Rafael Rafailov, Archit Sharma, Eric Mitchell, Christopher~D Manning, Stefano
  Ermon, and Chelsea Finn.
\newblock Direct preference optimization: Your language model is secretly a
  reward model.
\newblock In {\em NeurIPS}, 2023.

\bibitem{yang2025magma}
Jianwei Yang, Reuben Tan, Qianhui Wu, Ruijie Zheng, Baolin Peng, Yongyuan Liang,
  Yu~Gu, Mu~Cai, Seonghyeon Ye, Joel Jang, Yuquan Deng, Lars Liden, and Jianfeng
  Gao.
\newblock Magma: A foundation model for multimodal ai agents.
\newblock In {\em CVPR}, 2025.

\bibitem{li2024llavanext}
Bo~Li, Kaichen Zhang, Hao Zhang, Dong Guo, Renrui Zhang, Feng Li, Yuanhan Zhang,
  Ziwei Liu, and Chunyuan Li.
\newblock Llava-next: Stronger llms supercharge multimodal capabilities in the
  wild, 2024.
\newblock \url{https://llava-vl.github.io/blog/2024-05-10-llava-next-stronger-llms/}.

\bibitem{steiner2024paligemma2}
Andreas Steiner, Andr{\'e} Susano~Pinto, Michael Tschannen, Daniel Keysers,
  Xiao Wang, Yonatan Bitton, Alexey Gritsenko, Matthias Minderer, Anthony
  Sherbondy, Shangbang Long, Siyang Qin, Reeve Ingle, Emanuele Bugliarello,
  Sahar Kazemzadeh, Thomas Mesnard, Ibrahim Alabdulmohsin, Lucas Beyer, and
  Xiaohua Zhai.
\newblock Paligemma 2: A family of versatile vlms for transfer.
\newblock {\em arXiv preprint arXiv:2412.03555}, 2024.

\bibitem{wu2024deepseekvl2}
Zhiyu Wu, Xiaokang Chen, Zizheng Pan, Xingchao Liu, Wen Liu, Damai Dai, Huazuo
  Gao, Yiyang Ma, Chengyue Wu, Bingxuan Wang, Zhenda Xie, Yu~Wu, Kai Hu, Jiawei
  Wang, Yaofeng Sun, Yukun Li, Yishi Piao, Kang Guan, Aixin Liu, Xin Xie, Yuxiang
  You, Kai Dong, Xingkai Yu, Haowei Zhang, Liang Zhao, Yisong Wang, and Chong
  Ruan.
\newblock Deepseek-vl2: Mixture-of-experts vision-language models for advanced
  multimodal understanding.
\newblock {\em arXiv preprint arXiv:2412.10302}, 2024.

\bibitem{chen2024internvl25}
Zhe Chen, Weiyun Wang, Yue Cao, Yangzhou Liu, Zhangwei Gao, Erfei Cui, Jinguo
  Zhu, Shenglong Ye, Hao Tian, Zhaoyang Liu, et~al.
\newblock Expanding performance boundaries of open-source multimodal models with
  model, data, and test-time scaling.
\newblock {\em arXiv preprint arXiv:2412.05271}, 2024.

\bibitem{glm2025glm4v}
Z.ai.
\newblock Glm-4.6v: Open source multimodal models with native function calling.
\newblock Z.ai Technical Blog, 2025.

\bibitem{abouelenin2025phi4rv}
Jyoti Aneja et~al.
\newblock Phi-4-reasoning-vision-15b technical report.
\newblock {\em arXiv preprint arXiv:2603.03975}, 2026.

\end{thebibliography}

}
\end{document}